\documentclass[lettersize,journal]{IEEEtran}
\usepackage{amsmath,amsfonts}
\usepackage{algorithmic}
\usepackage{algorithm}
\usepackage{array}
\usepackage[caption=false,font=normalsize,labelfont=sf,textfont=sf]{subfig}
\usepackage{textcomp}
\usepackage{stfloats}
\usepackage{url}
\usepackage{verbatim}
\usepackage{graphicx}
\usepackage{cite}
\usepackage{xcolor}
\usepackage{hyperref}
\usepackage{bm}
\usepackage{booktabs}
\usepackage{multirow}
\usepackage{tabularx}
\usepackage{bbding}
\usepackage{amssymb}
\begin{document}
\newcommand{\fullmodelnames}{\textcolor{red}{Bayesian Class Adaptation plus }}
\newcommand{\shortmodelnames}{\textcolor{red}{BCA+ }}
\newcommand{\zhou}[1]{\textcolor{red}{#1}}

\title{Bayesian Test-time Adaptation for Object Recognition and Detection with Vision-language Models}

% LLM给的标题建议
% \title{BCA+: A Unified, Training-Free Framework for Bayesian Test-Time Adaptation in Vision-Language Models} 
% BCA+: Unified Bayesian Test-Time Adaptation for Object Detection and Recognition with Vision-Language Models


\author{Lihua Zhou,  
Mao Ye,~\IEEEmembership{Senior Member,~IEEE,}
Shuaifeng Li,
Nianxin Li,
Jinlin Wu,
Xiatian Zhu, \\
Lei Deng,
Hongbin Liu,
Jiebo Luo,~\IEEEmembership{Fellow,~IEEE,}
Zhen Lei$^*$, ~\IEEEmembership{Fellow,~IEEE} 
\thanks{
\IEEEcompsocthanksitem Lihua Zhou, Jinlin Wu and Hongbin Liu are with the Centre for Artificial Intelligence and Robotics, Hong Kong Institute of Science and Innovation, Chinese Academy of Sciences, Hong Kong, China.
Email: lihuazhou120@gmail.com, jinlin.wu@cair-cas.org.hk, liuhongbin@ia.ac.cn
\IEEEcompsocthanksitem Mao Ye, Shuaifeng Li and Nianxin Li are with School of Computer Science and Engineering, University of Electronic Science and Technology of China, Chengdu 611731, China. E-mail: maoye@uestc.edu.cn, hotwindlsf@gmail.com, linianxin1220@gmail.com
\IEEEcompsocthanksitem Xiatian Zhu is with Surrey Institute for People-Centred Artificial Intelligence, CVSSP, University of Surrey, Guildford, UK. E-mail: xiatian.zhu@surrey.ac.uk
\IEEEcompsocthanksitem {Lei Deng} is with the School of Electronics and Information Engineering, Shenzhen University, Shenzhen, China. E-mail: ldeng.sjtu@gmail.com
\IEEEcompsocthanksitem {Jiebo Luo} is with the University of Rochester and performed this work while on a sabbatical leave at the Hong Kong Institute of Science and Innovation.
\IEEEcompsocthanksitem Zhen Lei is with the School of Artificial Intelligence, University of Chinese Academy of Sciences (UCAS), Beijing 100049, China; the Centre for Artificial Intelligence and Robotics, Hong Kong Institute of
Science and Innovation, Chinese Academy of Sciences, Hong Kong, China.
Email: zhen.lei@ia.ac.cn
\IEEEcompsocthanksitem * corresponding author.
}
}

\markboth{Journal of \LaTeX\ Class Files,~Vol.~14, No.~8, August~2021}%
{Shell \MakeLowercase{\textit{et al.}}: A Sample Article Using IEEEtran.cls for IEEE Journals}


\maketitle

\begin{abstract}
Vision-language models (VLMs) such as CLIP and Grounding DINO have achieved remarkable success in object recognition and detection. However, their performance often degrades under real-world distribution shifts. Test-time adaptation (TTA) aims to mitigate this issue by adapting models during inference. Existing methods either rely on computationally expensive backpropagation, which hinders real-time deployment, or focus solely on likelihood adaptation, which overlooks the critical role of the prior.
Our prior work, Bayesian Class Adaptation (BCA), addressed these shortcomings for object recognition by introducing a training-free framework that incorporates adaptive priors.
Building upon this foundation, we now present Bayesian Class Adaptation plus (BCA+), a unified, training-free framework for TTA for both object recognition and detection. BCA+ introduces a dynamic cache that adaptively stores and updates class embeddings, spatial scales (for detection), and, crucially, adaptive class priors derived from historical predictions. We formulate adaptation as a Bayesian inference problem, where final predictions are generated by fusing the initial VLM output with a cache-based prediction. This cache-based prediction combines a dynamically updated likelihood (measuring feature and scale similarity) and a prior (reflecting the evolving class distribution). This dual-adaptation mechanism, coupled with uncertainty-guided fusion, enables BCA+ to correct both the model's semantic understanding and its contextual confidence. As a training-free method requiring no backpropagation, BCA+ is highly efficient. Extensive experiments demonstrate that BCA+ achieves state-of-the-art performance on both recognition and detection benchmarks.
\end{abstract}


\section{Introduction}

\IEEEPARstart{T}{he} ability to recognize and localize objects in images is a fundamental tasks in computer vision \cite{he2016deep,ren2016faster,deng2009imagenet,lin2014microsoft}. Recent advances in vision-language models (VLMs) \cite{zhang2024vision}, such as CLIP \cite{radford2021learning} and Grounding DINO \cite{liu2024grounding}, have revolutionized these tasks by leveraging large-scale image-text pairs to learn robust, multi-modal representations. These models achieve remarkable zero-shot performance, enabling object recognition and detection without the need to train a dedicated baseline model from scratch for each new task.

\begin{figure}[t]
  \begin{center}
  \includegraphics[width=0.4\textwidth]{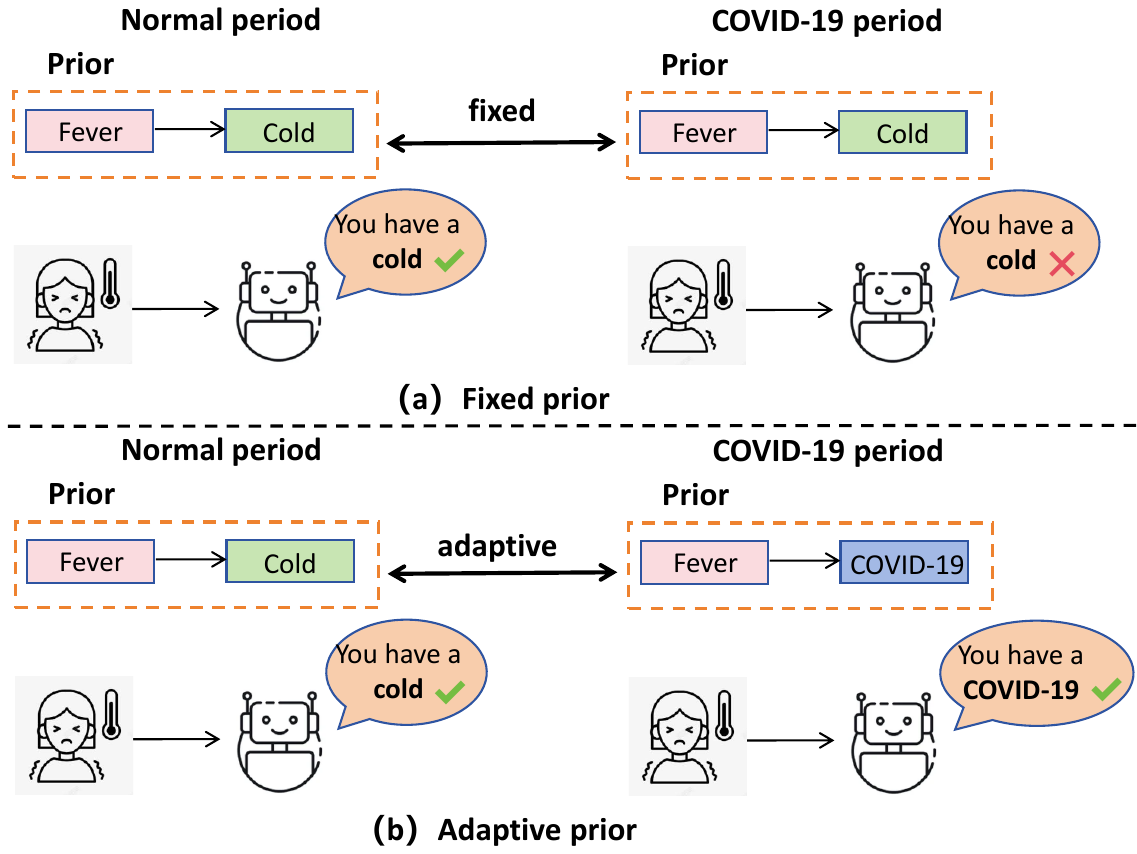}
  \end{center}
     \caption{Illustrative example of Fixed Prior \textit{vs.} Adaptive Prior: Comparison of Diagnosis Outcomes. In the fixed prior scenario, patients with fever are consistently diagnosed with the common cold, regardless of whether it is a normal period or a COVID-19 period. In contrast, the adaptive prior scenario adjusts the diagnosis based on the current context. During normal periods, patients with fever are diagnosed with the common cold, while during the COVID-19 period, they are more likely to be diagnosed with COVID-19. This demonstrates the importance of performing prior adaptation in different environments.
}
  \label{fig:1}
\end{figure}

Despite their impressive capabilities, VLMs are often brittle when deployed in real-world scenarios \cite{zhou2022learning,zhou2022conditional,gao2024clip,zhang2021tip}. This brittleness stems primarily from the distribution shift between the training and test data \cite{qu2025glc++,pei2025selection}. While VLMs are pre-trained on large-scale, general-purpose datasets, real-world test data often comes from specific domains with distinct characteristics, such as varying environmental conditions, object appearances, or background clutter \cite{li2022source,li2024cloud}. This discrepancy between the broad, general distribution of the pre-training data and the narrow, domain-specific distribution of the test data leads to significant performance degradation \cite{li2024cloud,li2022source}.
To address this challenge, the paradigm of Test-Time Adaptation (TTA) has emerged \cite{liang2024comprehensive}. TTA aims to adapt pre-trained models online to the test data distribution during inference, without requiring access to the original training data or labels. By continuously updating the model based on incoming test samples, TTA enhances the model's robustness and adaptability to new environments. Crucially, an effective TTA method must not only improve accuracy but also preserve inference efficiency, ensuring that the adaptation process is fast and lightweight enough for real-time deployment.

For object recognition, numerous methods based on VLMs like CLIP \cite{radford2021learning} have been proposed \cite{karmanov2024efficient,han2024dota,feng2023diverse,shu2022test} and can be broadly divided into two categories. The first relies on fine-tuning text prompts via backpropagation to minimize prediction entropy \cite{feng2023diverse, shu2022test}. While effective, this approach is computationally expensive and incompatible with the real-time requirement of TTA. The second category adopts a training-free strategy, using a memory bank to store historical visual embeddings and updating class embeddings statistically \cite{karmanov2024efficient, han2024dota}. These methods enable efficient, real-time adaptation. However, from a Bayesian perspective, these methods focus solely on likelihood adaptation, updating class embeddings, while neglecting the prior, which reflects the model’s belief about the current class distribution. Since the posterior depends on both likelihood and prior, ignoring prior adaptation limits robustness, especially when the test-time class distribution shifts significantly. We illustrate the importance of prior adaptation using an example in Figure~\ref{fig:1}, showing how the optimal class distribution varies drastically across different test environments.

\begin{figure}[t]
    \centering
    \includegraphics[width=0.5\textwidth]{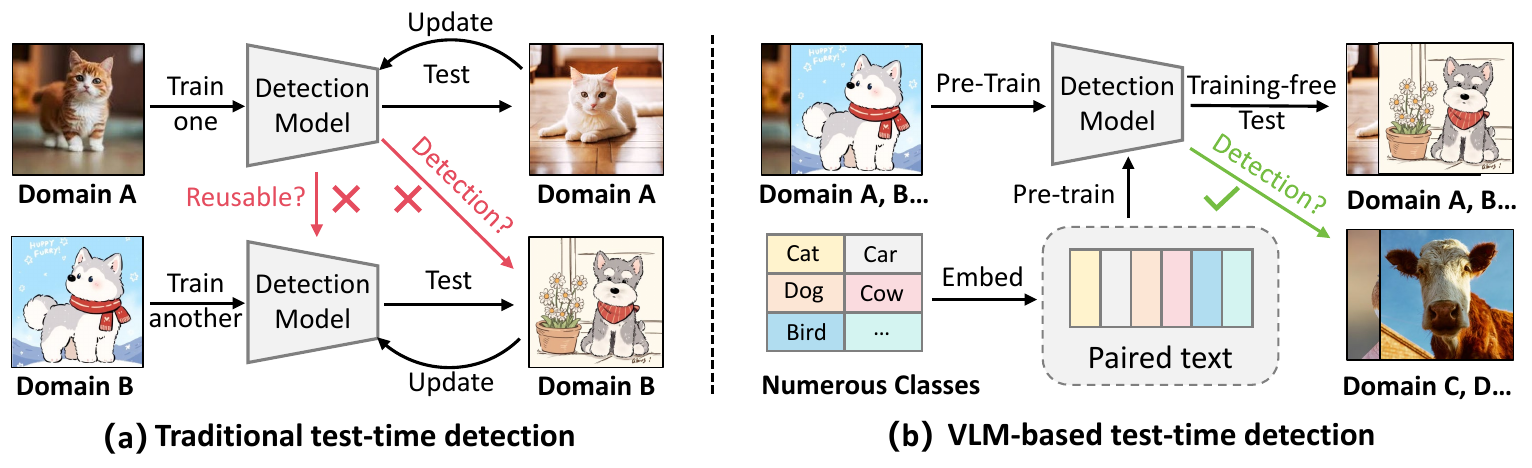}
    \caption{Traditional test-time detection \textit{vs.} VLM-based test-time detection. {Traditional test-time detection requires training a dedicated detector (e.g., Faster R-CNN) for each specific domain. For example, a model trained on the "Cat" domain cannot be directly applied to the "Dog" domain and must be retrained from scratch, which is resource-intensive. Even during test-time adaptation, these methods rely on backpropagation for updates. While VLM-based method uses a single VLM (e.g., Grounding DINO) as a universal baseline, enabling detection for any category (e.g., both "Cat" and "Dog") without retraining. Our BCA+ framework enables training-free adaptation via a dynamic cache, eliminating backpropagation for real-time deployment.}}
    \label{fig:setting}
\end{figure}

In contrast, for object detection, the landscape is markedly different. Despite the success of VLMs in TTA for recognition, there is a striking lack of work on TTA for object detection using VLMs like Grounding DINO \cite{liu2024grounding}. Existing TTA methods for detection \cite{chen2023stfar,yoo2024and} still use traditional detectors like Faster R-CNN \cite{ren2016faster}. A major drawback of these methods is that they require training a new model from scratch for each specific task or domain. This makes them very costly in terms of time and resources. In contrast, VLMs like Grounding DINO \cite{liu2024grounding} can naturally handle open-vocabulary detection, meaning they can detect objects from any category defined by text, without needing to retrain a new model. This makes them a much more flexible and scalable choice. Moreover, these existing TTA methods for detection usually update the model during testing by using backpropagation. This process is very slow and takes a long time, which goes against the main goal of TTA: to adapt quickly and efficiently in real time. Therefore, there is a strong need for a new TTA method that is both training-free and can work effectively with powerful VLMs like Grounding DINO for the object detection task.

To address these challenges, we propose Bayesian Class Adaptation plus (BCA+), a unified, training-free framework for test-time adaptation in both object recognition and detection. 
The process unfolds in three key stages. First, for each incoming image, the pre-trained VLM generates an initial prediction along with its visual features (and bounding box for detection). These initial predictions can be misled by distribution shifts. Second, to correct these predictions, BCA+ employs a dynamic cache that stores historical information. We formulate a cache-based prediction by treating the adaptation process as a Bayesian inference problem. This cache-based prediction is derived by combining a likelihood, which measures how similar the current image is to previously seen patterns (based on visual features and spatial scales), with a dynamically updated prior, which captures the belief about which categories are more likely to appear in the current environment. Third, BCA+ performs an uncertainty-guided fusion to produce the final prediction. This fusion intelligently combines the initial VLM prediction and the cache-based prediction, giving more weight to the one that is more confident. Finally, the cache itself is continuously updated: high-confidence final predictions are used to either create new entries for novel patterns or to incrementally refine existing entries in the cache, ensuring the model's knowledge grows and adapts over time. This entire process is training-free, enabling BCA+ to achieve robust and accurate predictions for both tasks while maintaining high inference efficiency.

BCA+ extends our prior framework, Bayesian Class Adaptation (BCA) \cite{zhou2025bayesian}, in three key areas: research scope, methodology, and evaluation. Conceptually, BCA+ broadens its application from exclusively object recognition to a unified framework that seamlessly handles both recognition and the more complex task of object detection. Methodologically, we introduce a dynamic cache that adaptively manages historical information, including class embeddings, spatial scales, and priors, departing from BCA’s fixed cache. This is complemented by a new dual-stream prediction framework and an uncertainty-guided fusion strategy that intelligently combines predictions from both the pre-trained VLM and our dynamic cache, leading to more robust results. Empirically, we provide a more comprehensive validation, including extensive new benchmarks for object detection and in-depth analyses, demonstrating that BCA+ achieves state-of-the-art performance with high efficiency.
Our key contributions are as follows:
\begin{itemize}
    \item We present BCA+, a unified, training-free TTA framework that effectively handles both object recognition and object detection. To the best of our knowledge, this is the first work to apply TTA to object detection using VLMs.
    \item We introduce a dynamic cache that simultaneously adapts both the likelihood (class embeddings and spatial scales for detection) and the prior (class distribution). This dual-adaptation mechanism provides a more comprehensive and robust solution to handle distribution shifts.
    \item We introduce a novel dual-stream prediction framework and an uncertainty-guided fusion strategy. This methodological advance, which is not present in BCA, intelligently combines predictions from the pre-trained VLM and our dynamic cache, leading to more robust results.
    \item We demonstrate through extensive experiments that BCA+ achieves state-of-the-art performance on standard benchmarks for both tasks, while maintaining high inference efficiency suitable for real-time applications.
\end{itemize}

% The rest of this paper is organized as follows. Section~\ref{sec:related} reviews related work in vision-language models and test-time adaptation. Section~\ref{sec:pre} introduces the preliminaries, with a focus on the prediction mechanisms of CLIP and Grounding DINO. Section~\ref{sec:method} presents the proposed BCA+ framework in detail. Section~\ref{sec:exp} describes the experimental setup and reports the results. Finally, Section~\ref{sec:conclusion} concludes the paper.

\section{Related Work}\label{sec:related} 

\subsection{Vision-Language Models}
Traditional computer vision systems often require training a dedicated deep neural network for each specific task, a process that is both labor-intensive and lacks generalization \cite{he2016deep,ren2016faster,deng2009imagenet,lin2014microsoft}. This paradigm has been challenged by the emergence of VLMs \cite{zhang2024vision}. By learning  multi-modal representations from massive web-scale image-text pairs, VLMs can perform a wide range of visual tasks with a single, pre-trained model, enabling zero-shot and open-vocabulary capabilities.

A seminal work in this field is CLIP \cite{radford2021learning}, which uses contrastive learning to align image and text embeddings in a shared semantic space. This alignment allows CLIP to perform zero-shot object recognition: given a set of class names, the model can classify an image by comparing its visual embedding to the text embeddings of the class names, without any task-specific fine-tuning. The success of CLIP has inspired a wave of follow-up research. ALIGN \cite{jia2021scaling} demonstrated the power of scale by training on over a billion noisy image-text pairs. FILIP \cite{yaofilip} introduced a fine-grained interaction mechanism to better capture local semantic alignments between images and text. PaLI \cite{chenpali} explored joint scaling of both the vision and language components, pushing the boundaries of model size and capability. ZeroVL \cite{cui2022contrastive} focuses on data-efficient pre-training by introducing a novel data augmentation strategy into the contrastive learning framework. LiT \cite{zhai2022lit} proposes contrastive tuning with a locked image encoder, fine-tuning only a lightweight text-side adapter for efficient knowledge transfer.

The success of VLMs in object recognition has inspired a range of approaches to extend their capabilities to the more complex task of object detection. This has led to two main research directions: knowledge distillation and end-to-end open-vocabulary detectors.
The first direction focuses on distilling knowledge from powerful VLMs like CLIP into existing detector frameworks. 
ViLD \cite{guopen} is a seminal work in this area, which distills the image-level knowledge from a CLIP teacher into a two-stage detector, enabling it to detect objects from arbitrary categories. Another approach leverages VLMs to generate pseudo-labels for training. 
DetCLIP \cite{yao2022detclip} uses a large-scale image captioning dataset to generate pseudo bounding box labels, effectively expanding the detector's knowledge base and generalization ability.
The second, and more recent, direction aims to build end-to-end open-vocabulary detectors that natively integrate VLMs. 
OV-DETR \cite{zareian2021open} uses image and text embeddings from a CLIP model as learnable queries within the DETR framework \cite{carion2020end}, allowing it to decode category-specified boxes directly from language inputs. 
GLIP \cite{li2022grounded} formulates object detection as a visual grounding problem, leveraging additional phrase-grounding data to learn aligned semantics at both the phrase and region levels. This formulation has proven so effective that GLIP can even surpass traditional detectors on fully-supervised benchmarks. Building upon these ideas, Grounding DINO \cite{liu2024grounding} represents a major leap forward by integrating a language-guided query selection mechanism and a feature enhancement module, achieving state-of-the-art performance in open-vocabulary detection.

Our work builds upon these powerful VLMs, using CLIP \cite{radford2021learning} for object recognition and Grounding DINO \cite{liu2024grounding} for object detection as our base models.

\subsection{Test-Time Adaptation}
Test-Time Adaptation aims to improve the performance of a pre-trained model on distribution-shifted test data by adapting it online, without access to the original training data or labels. Early TTA methods were primarily developed for standard deep neural networks and can be applied to both object recognition and object detection. For recognition, common strategies included entropy minimization \cite{mirza2022norm,zhaodelta}, recalibration of batch normalization statistics \cite{niutowards,wangtent}, pseudo-labeling \cite{iwasawa2021test,jangtest} and consistency regularization \cite{wang2022continual,niu2022efficient}. For object detection, methods like STFAR \cite{chen2023stfar} mitigated the impact of noisy pseudo-labels through feature distribution regularization, while MemCLR \cite{vs2023towards} enhanced target-specific representations using contrastive learning in dynamic environments. Some approaches \cite{yoo2024and} also developed online frameworks with dynamic parameter optimization to handle continuous domain shifts. While these early methods demonstrated effectiveness, they are often tightly coupled to specific model architectures and require careful hyperparameter tuning. More importantly, they typically rely on backpropagation for adaptation, which is computationally expensive and hinders real-time deployment. These limitations highlight the need for a more flexible and efficient adaptation paradigm.

The advent of powerful pre-trained VLMs like CLIP has opened a new frontier for TTA. These models, with their inherent zero-shot capabilities, provide a flexible and generalizable foundation for adaptation. Benefiting from the success of CLIP, a growing body of work has focused on test-time adaptation for object recognition. Early approaches like TPT \cite{shu2022test} optimize text prompts by minimizing prediction entropy across augmented views, while DiffTPT \cite{feng2023diverse} and C-TPT \cite{yoonc} enhance this process with better data augmentation and prompt calibration. PromptAlign \cite{samadhalign} aligns the statistics of test samples with those of the source domain by matching their token distribution, thereby optimizing the text prompt for improved generalization.
HisTPT \cite{zhang2024historical} introduces multiple knowledge banks to maintain a comprehensive memory of past data, effectively mitigating catastrophic forgetting during test-time adaptation.
However, these methods rely on backpropagation during testing, which is computationally costly.

To achieve real-time adaptation, a new line of research has emerged, focusing on training-free methods that avoid backpropagation. These approaches adapt the model using only forward passes and simple updates. TDA \cite{karmanov2024efficient} uses a lightweight cache to progressively refine pseudo-labels. DOTA \cite{han2024dota} continually estimates the distribution of test samples. MTA \cite{zanella2024test} manages augmented views in a training-free manner. Zero \cite{farina2024frustratingly} proposes a simple yet effective strategy by setting the Softmax temperature to zero, converting soft probability distributions into hard one-hot encodings to guide model adaptation.

Despite all this progress in recognition,
there are very few methods that apply TTA to open-vocabulary object detection using VLMs like Grounding DINO. Most existing TTA methods for detection are built on older, closed-set models like Faster RCNN and still rely on slow, gradient-based updates. To fill this gap, we propose BCA+, a unified, training-free framework that can adapt both CLIP for recognition and Grounding DINO for detection in real time.

\section{Preliminaries}\label{sec:pre}

\noindent\textbf{CLIP.} CLIP \cite{radford2021learning} is a vision-language model designed for zero-shot object recognition tasks, leveraging large-scale paired image-text data to learn robust, multi-modal representations. Trained via contrastive learning, CLIP aligns images and their corresponding textual descriptions in a shared embedding space, enabling generalization across diverse visual recognition tasks without task-specific fine-tuning.

CLIP comprises two main components: a visual encoder \(E_v\) (e.g., ResNet-50 \cite{he2016deep} or ViT-B/16 \cite{dosovitskiy2020image}) and a text encoder \(E_t\) (e.g., a Transformer-based model \cite{vaswani2017attention}). For a \(K\)-class recognition task, CLIP processes a set of \(K\) hand-crafted text prompts, such as ``a photo of [class \(k\)]'' for \(k=1,\dots,K\). These prompts are fed into the text encoder \(E_t\), which generates normalized text embeddings \(\boldsymbol{F}_t = \{ \bm{f}^t_k \}_{k=1}^K \in \mathbb{R}^{K \times d}\), where \(d\) is the embedding dimension. For an input test image \(\bm{x}_i\), the visual encoder \(E_v\) extracts a normalized visual embedding \(\bm{f}^v_i \in \mathbb{R}^d\), capturing the semantic content of the image.
To compute predictions, CLIP measures the cosine similarity between the visual embedding \(\bm{f}^v_i\) and each text embedding \(\bm{f}^t_k\). The probability distribution over the \(K\) classes is then calculated as:
\begin{equation}
    % P(Y|\bm{x}_i) = {\text{Softmax}(\cos(\bm{f}_{i}^v, \bm{F}_t))}
    P_{clip}(Y=k|\bm{x}_i) = \frac{\text{exp}(\cos(\bm{f}_{i}^v, \bm{f}_k^t)))}{\sum_{j=1}^K \text{exp}(\cos(\bm{f}_{i}^v, \bm{f}_j^t)))},
    \label{Eq:01}
\end{equation}
where \(\cos(\cdot, \cdot)\) denotes the cosine similarity, and \(\text{exp}(\cdot)\) denotes the exponential function. 

\noindent \textbf{Grounding DINO.} Grounding DINO \cite{liu2024grounding} is a state-of-the-art vision-language model pivotal for open-vocabulary object detection, leveraging large-scale image-text pairs to align visual and textual modalities in a shared embedding space \cite{li2024cloud}. Trained with detection and grounding objectives, it enables robust detection across arbitrary categories specified by text prompts, serving as a powerful backbone for test-time adaptive object detection.

Grounding DINO employs a dual-encoder architecture comprising a visual encoder \(E_v\) (e.g., Swin Transformer \cite{liu2021swin}) and a text encoder \(E_t\) (e.g., BERT-based \cite{devlin2019bert}). For an input image \(\bm{x}_i\) and a set of text prompts \(\mathbf{T} = \{[t_k]\}_{k=1}^K\), where \([t_k]\) denotes the \(k\)-th category (e.g., ``cat . dog .''), the text encoder \(E_t\) processes \(\mathbf{T}\) to produce vanilla text embeddings \(\bm{F}_t' = E_t(\mathbf{T}) \in \mathbb{R}^{K \times d}\). The visual encoder \(E_v\) extracts vanilla vision features \(\bm{F}_v' = E_v(\bm{x}_i)\), which are then fused with \(\bm{F}_t'\) via an attention-based feature enhancer, yielding enhanced visual features \(\bm{F}_v = \psi(\bm{F}_v', \bm{F}_t') \in \mathbb{R}^{N_I \times d}\) and enhanced text features \(\bm{F}_t = \psi(\bm{F}_t', \bm{F}_v')= \{ \bm{f}^t_k \}_{k=1}^K \in \mathbb{R}^{K \times d}\), where \(N_I\) is the number of image tokens.

To generate detection outputs, Grounding DINO selects \(N = 900\) cross-modality queries \(\{\bm{f}_{ij}'\}_{j=1}^N\) from \(\bm{F}_v\) using a language-guided query selection mechanism based on similarity with \(\bm{F}_t\). These queries are refined through a DETR-like \cite{zhangdino} cross-modality decoder, producing updated query features \(\{\bm{f}_{ij}^v\}_{j=1}^N\), where each \(\bm{f}_{ij}^v \in \mathbb{R}^d\) serves as a proposal feature. Then class probabilities for each proposal are computed as:
\begin{equation}
P_{gdino}(Y=k|\bm{x}_i) = \frac{\text{exp}(\sigma(\cos(\bm{f}_{ij}^v, \bm{f}_k^t))))}{\sum_{l=1}^K \text{exp}(\sigma(\cos(\bm{f}_{ij}^v, \bm{f}_l^t))))},
\label{Eq:02}
\end{equation}
where \(P(Y|\bm{x}_{ij})\) denotes the prediction of $j$-th proposal of \(\bm{x}_i\), \(\sigma(\cdot)\) is the sigmoid function. For bounding box prediction, the decoder iteratively refines positional information across its layers, and a regression head processes the final refined proposal feature, combining it with the previous layer’s prediction, to produce coordinates:
\begin{equation}
\bm{b}_{ij} = \bm{b}_{ij}^{-1} + \rho(\bm{f}_{ij}^v),
\label{Eq:03}
\end{equation}
where \(\bm{b}_{ij} \in \mathbb{R}^4\) represents the bounding box \([x, y, w, h]\) with center coordinates \(x, y\) and width-height \(w, h\), \(\bm{b}_{ij}^{-1} \in \mathbb{R}^4\) is the bounding box prediction from the previous decoder layer, and \(\rho\) is a regression head applied to the final \(\bm{f}_{ij}^v\) after iterative refinement.

\section{Method}\label{sec:method}

\noindent \textbf{Problem Setting.} 
In this work, we address the challenge of test-time adaptation for pre-trained vision-language models, focusing on both object recognition and detection. Given a sequence of test images $\{\bm{x}_i\}_{i=1}^n$ that arrive sequentially, our goal is to adapt a pre-trained model to predict accurate outputs for each image $\bm{x}_i$ immediately in an online manner, despite potential distribution shifts between the pre-training data and test data. For object recognition, we leverage CLIP \cite{radford2021learning} to generate initial predictions for $K$ predefined categories. For object detection, we employ Grounding DINO \cite{liu2024grounding} to predict a set of bounding boxes and corresponding initial class probabilities among $K$ predefined categories. While these models provide strong initial predictions, their performance can still degrade under real-world distribution shifts. To address this issue, our method introduces a Bayesian framework that leverages a dynamically updated cache to refine the initial predictions, thereby improving the model's robustness and accuracy, while preserving inference efficiency for real-time applications.

\noindent \textbf{Notation Unification.}
To unify the notation for both tasks, we treat the recognition task as a special case of detection where there is only one proposal per image. Specifically, the visual embedding $\bm{f}_i^v$ of CLIP is redefined as $\bm{f}_{ij}^v$ (where $j = 1$) to align with the detection task's notation. Formally, when an image $\bm{x}_i$ is processed by the vision-language model, the outputs can be represented as follows: for object recognition, we obtain $\{(\bm{f}_{ij}^v, \bm{p}^{init}_{ij})\}_{j=1}^1$, where $\bm{f}_{ij}^v$ is the visual embedding of the single proposal and $\bm{p}^{init}_{ij} = P_{clip}(Y|\bm{x}_i)$ is the initial class probability computed using Eq.~\eqref{Eq:01}; for object detection, we obtain $\{(\bm{f}_{ij}^v, \bm{b}_{ij}, \bm{p}^{init}_{ij})\}_{j=1}^N$, where $\bm{f}_{ij}^v$ is the visual embedding of the $j$-th proposal, $\bm{b}_{ij}$ is the predicted bounding box, and $\bm{p}^{init}_{ij} = P_{gdino}(Y|\bm{x}_{ij})$ is the initial class probability computed using Eq.~\eqref{Eq:02}. 

\begin{figure*}[t]
    \centering
    \includegraphics[width=\textwidth]{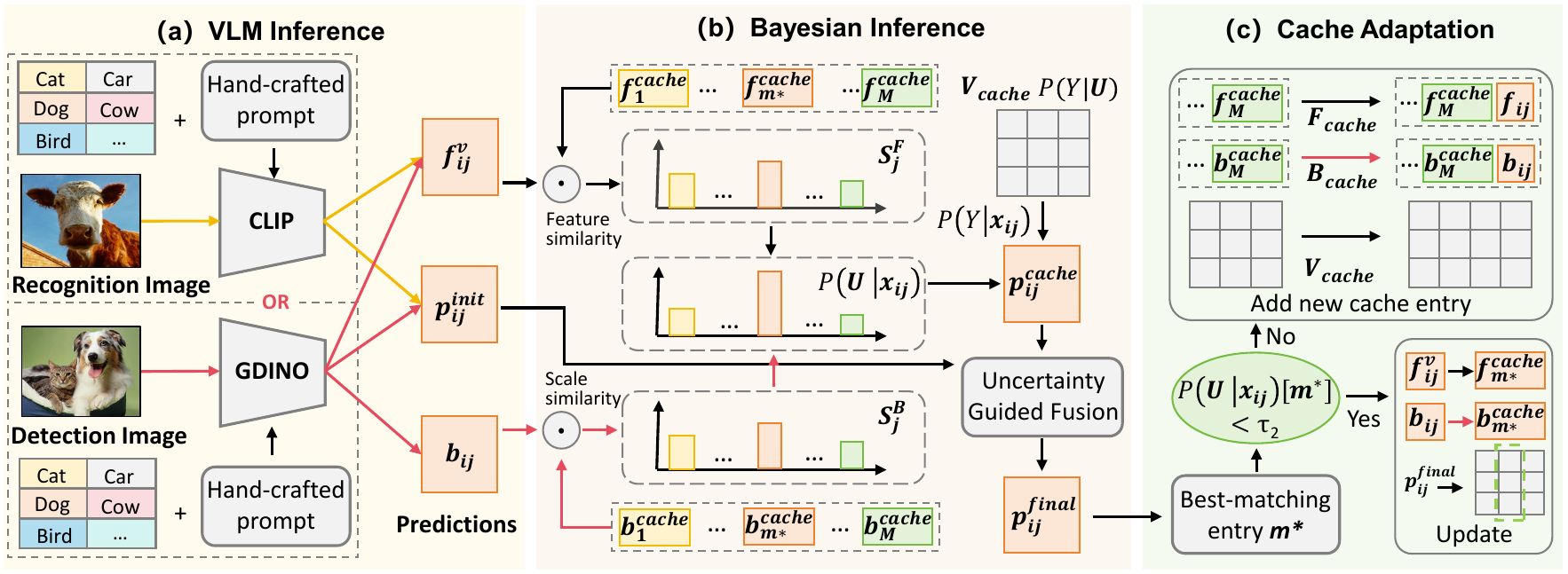}
    \caption{Overview of the proposed BCA+ framework. The process is divided into three main stages: (a) \textbf{VLM Inference}, where the pre-trained vision-language model (CLIP for recognition or Grounding DINO for detection) processes the input image $\bm{x}_i$ to generate initial outputs, including visual embeddings $\{\bm{f}_{ij}^v\}_j$, initial class predictions $\{\bm{p}^{init}_{ij}\}_j$, and bounding boxes $\{\bm{b}_{ij}\}_j$ (for detection). (b) \textbf{Bayesian Inference}, which computes a cache-based prediction $\{\bm{p}^{cache}_{ij}\}_j$ by combining a likelihood (derived from feature and scale similarity with cached entries) and a dynamically updated prior. The final prediction $\{\bm{p}^{final}_{ij}\}_j$ is produced by fusing the initial and cache-based predictions with an uncertainty-guided strategy. (c) \textbf{Cache Adaptation}, where the cache is updated based on the final prediction: likelihood adaptation refines the cached feature embeddings $\{\bm{f}^{cache}_m\}_m$ and scales $\{\bm{b}^{cache}_m\}_m$, while prior adaptation updates the cached priors $\{\bm{v}^{cache}_m\}_m$. Orange lines denote components exclusive to object recognition, red lines denote components exclusive to object detection, and black lines denote components shared by both tasks.}
    \label{fig:overview}
\end{figure*}

\noindent \textbf{Overview.} 
Our proposed Bayesian Class Adaptation plus (BCA+) is a training-free method for test-time adaptation in both object recognition and detection tasks, extending the original BCA framework to handle distribution shifts robustly during inference. As illustrated in Fig.~\ref{fig:overview}, the BCA+ framework operates in a closed-loop process consisting of three key stages.
{(a) VLM Inference:} An input image $\bm{x}_i$ is first processed by a pre-trained vision-language model (CLIP for recognition, Grounding DINO for detection). This generates the initial outputs: visual embeddings $\{\bm{f}_{ij}^v\}_j$, initial class predictions $\{\bm{p}^{init}_{ij}\}_j$, and bounding boxes $\{\bm{b}_{ij}\}_j$ (for detection).
{(b) Bayesian Inference:} A cache-based prediction $\{\bm{p}^{cache}_{ij}\}_j$ is computed by formulating the adaptation as a Bayesian inference problem. This involves calculating a likelihood based on the similarity between the current proposal's features (and scales for detection) and those stored in the cache, and combining it with a dynamically updated prior. The final prediction $\{\bm{p}^{final}_{ij}\}_j$ is then obtained by fusing the initial prediction $\{\bm{p}^{init}_{ij}\}_j$ and the cache-based prediction $\{\bm{p}^{cache}_{ij}\}_j$ using an uncertainty-guided fusion strategy.
{(c) Cache Adaptation:} The cache is dynamically updated based on high-confidence final predictions. This update has two components: \textit{likelihood adaptation}, which refines the cached feature embeddings $\{\bm{f}^{cache}_m\}_m$ and scales $\{\bm{b}^{cache}_m\}_m$ using statistical aggregation, and \textit{prior adaptation}, which updates the cached priors $\{\bm{v}^{cache}_m\}_m$ to reflect the current test-time class distribution.
By iteratively performing these three stages for each incoming test sample, BCA+ achieves robust and accurate predictions for both tasks under distribution shifts.

\subsection{Bayesian Inference for Cache-Based Predictions}

As illustrated in the Bayesian Inference stage of Figure~\ref{fig:overview}, our method refines the initial predictions $\{\bm{p}^{init}_{ij}\}_j$ from the pre-trained VLM (detailed in Section~\ref{sec:pre}) by leveraging a dynamic cache. This section details the core of our framework: how we formulate this refinement as a Bayesian inference problem to compute a cache-based prediction $\{\bm{p}^{cache}_{ij}\}_j$.
% Despite the robust predictions of VLMs like Grounding DINO for object detection \cite{liu2024grounding} and CLIP for object recognition \cite{radford2021learning}, distribution shifts between training and test data can still degrade their performance when deployed in real-world environments. To address this, we employ a cache to store cached class embeddings, cached scales (for detection), and cached class priors, enabling the correction of predictions to bridge the train-test distribution gap. Specifically, for each incoming image $\bm{x}_i$, we leverage the cache based on Bayesian estimation to refine the initial predictions \(\{\bm{p}^{init}_{ij}\}_j\) generated by the vision-language model. This refinement is achieved with the help of visual embeddings \(\{\bm{f}^{v}_{ij}\}_j\) (and bounding boxes \(\{\bm{b}_{ij}\}_j\) for detection), which are used to compute the cache-based predictions \(\{\bm{p}^{cache}_{ij}\}_j\) that align the pre-trained VLM's outputs with the test-time distribution. The BCA+ framework leverages this cache to dynamically adapt to distribution shifts while maintaining computational efficiency.

\subsubsection{\textbf{Analysis}}
In this section, we analyze how our method generates cache-based predictions $\{\bm{p}^{cache}_{ij}\}_j$ based on cached historical information \(\{\bm{\mu}_{m}\}_{m=1}^M\) for both object recognition and object detection tasks. For recognition, the input is a visual embedding $\bm{x}_{ij} = (\bm{f}_{ij}^v)$ and $\bm{\mu}_m = (\bm{f}_m^{cache})$, where $\bm{f}_m^{cache} \in \mathbb{R}^d$ represents the averaged class embedding derived from historical visual embedding. For detection, the input consists of high-confidence proposals $\bm{x}_{ij} = (\bm{f}_{ij}^v, \bm{b}_{ij})$ and $\bm{\mu}_m = (\bm{f}_m^{cache}, \bm{b}_m^{cache})$, where $\bm{b}_m^{cache} \in \mathbb{R}^2$ stores the average spatial scale \([w, h]\).
Importantly, the number of cached entries $M$ is not fixed and can be dynamically expanded during inference. This flexibility allows our model to incorporate new knowledge on-the-fly, such as storing additional information in response to novel objects or spatial patterns observed during test time. By leveraging the cached historical information $\{\bm{\mu}_m\}_{m=1}^M$, we compute the posterior probability of class label $Y$ given input $\bm{x}_{ij}$ as follows:
\begin{equation}
\begin{aligned}
    P(Y|\bm{x}_{ij}) &=\sum_{m=1}^M P(Y,\bm{\mu}_m|\bm{x}_{ij}) \\
    &=\sum_{m=1}^M  P(\bm{\mu}_m|\bm{x}_{ij}) * P(Y|\bm{x}_{ij},\bm{\mu}_m) \\
    &= \sum_{m=1}^M P(\bm{\mu}_m|\bm{x}_{ij})* P(Y|\bm{\mu}_m).
    \label{Eq:1}
    \end{aligned}
\end{equation}
The derivation of the posterior probability follows a standard Bayesian treatment. The first line in Eq.~\eqref{Eq:1} is an application of the law of total probability, summing over all \(\bm{\mu}_m\). The second line applies the definition of conditional probability. The third line introduces a key assumption: once \(\bm{\mu}_m\) is known, the input \(\bm{x}_{ij}\) provides no additional information about the class label \(Y\) \cite{kruschke2010bayesian}. Further, applying Bayesian theorem, the posterior probability \(P(\bm{\mu}_m|\bm{x}_{ij})\) can be further written as:
\begin{equation}
\begin{aligned}
    P(\bm{\mu}_m|\bm{x}_{ij}) &= \frac{P(\bm{x}_{ij}|\bm{\mu}_m)*P(\bm{\mu}_m)}{P(\bm{x}_{ij})} \\
    &=\frac{P(\bm{x}_{ij}|\bm{\mu}_m)*P(\bm{\mu}_m)}{\sum_{j=1}^M P(\bm{x}_{ij}, \bm{\mu}_j)}\\
    &=\frac{P(\bm{x}_{ij}|\bm{\mu}_m)*P(\bm{\mu}_m)}{\sum_{j=1}^M P(\bm{x}_{ij}| \bm{\mu}_j) * P(\bm{\mu}_j)}
    \label{Eq:2}
\end{aligned}
\end{equation}
where $P(\bm{x}_{ij}|\bm{\mu}_m)$ represents likelihood, and $P(\bm{\mu}_m)$ is the probability of each cached information. Generally, it is assumed a uniform prior over all \(\bm{\mu}_m\), i.e. $P(\bm{\mu}_m)=\frac{1}{M}$, reflecting no preference in the absence of additional information.
By substituting Eq. \eqref{Eq:2} and $P(\bm{\mu}_m)=\frac{1}{M}$ back into Eq. \eqref{Eq:1}, we can obtain:
\begin{equation}
\begin{aligned}
    P(Y|\bm{x}_{ij}) &= \sum_{m=1}^M \frac{ P(\bm{x}_{ij}|\bm{\mu}_m)}{\sum_{j=1}^M P(\bm{x}_{ij}| \bm{\mu}_j)}* P(Y|\bm{\mu}_m). 
    \label{Eq:3}
\end{aligned}
\end{equation}

\textbf{Remark:} \textbf{(1): Connection to VLMs:}
The formulation in Eq.~\eqref{Eq:3} is general and can be applied to vision-language models, including CLIP \cite{radford2021learning} and Grounding DINO \cite{liu2024grounding}. In both models, the number of class embeddings is set as $M=K$, and each class embedding is defined as $\bm{\mu}_k = \bm{f}_{k}^t$. For the likelihood $P(\bm{x}_{ij}|\bm{\mu}_k)$, CLIP uses cosine similarity between the visual embedding $\bm{f}_{ij}^v$ and cached embedding $\bm{\mu}_k$: $P(\bm{x}_{ij}|\bm{\mu}_k) \propto \exp(\cos(\bm{f}_{ij}^v, \bm{\mu}_k))$. While Grounding DINO enhances this by applying a sigmoid activation function $\sigma(\cdot)$ to modulate the similarity scores: $P(\bm{x}_{ij}|\bm{\mu}_k) \propto \exp(\sigma(\cos(\bm{f}_{ij}^v, \bm{\mu}_k)))$. For the prior $P(Y|\bm{\mu}_k) \in \mathbb{R}^K$, both CLIP and Grounding DINO initially use a fixed one-hot vector derived from labels, where the $k$-th element is 1 and all other are 0. This means that when a sample belongs to the class embeddings $\bm{\mu}_k$, its probability of being in $k$-th category is 1. By substituting these into Eq. \eqref{Eq:3}, we can obtain Eq. \eqref{Eq:01} for CLIP and Eq. \eqref{Eq:02} for Grounding DINO.
\textbf{(2): Core Factors and Our Solution:}  From Eq.~\eqref{Eq:3}, we observe that computing the prediction \( P(Y|\bm{x}_{ij})\) hinges on two components: the likelihood \(P(\bm{x}_{ij}|\bm{\mu}_m)\), which quantifies the alignment between the current proposal and cached entries \(\bm{\mu}_m\), and the prior \(P(Y|\bm{\mu}_m)\), which captures the class distribution for each \(\bm{\mu}_m\). However, in VLMs, both $\bm{\mu}_m$ and \(P(Y|\bm{\mu}_m)\) are fixed during inference. These fixed parameters are typically derived from pre-trained data, which may lead to a significant distribution shift when applied to test data with different characteristics. Such discrepancies can degrade model performance, particularly in open-vocabulary or cross-domain scenarios. To address this issue, we propose a dynamic caching mechanism that continuously updates both the likelihood and the prior based on incoming test data. By leveraging this dynamic cache-based approach, our method effectively bridges the train-test distribution gap, significantly improving model robustness and accuracy across diverse scenarios.

\subsubsection{\textbf{Implementation}} 
For the current \(i\)-th image \(\bm{x}_i\), the pre-trained VLM extracts $\{(\bm{f}_{ij}^v, \bm{p}^{init}_{ij})\}_{j=1}^1$ for recognition or high-confidence proposals $\{(\bm{f}_{ij}^v, \bm{b}_{ij}, \bm{p}^{init}_{ij})\}_{j=1}^N$ for detection as mentioned earlier, which makes $\bm{x}_{ij} = (\bm{f}_{ij}^v)$ or $\bm{x}_{ij} = (\bm{f}_{ij}^v, \bm{b}_{ij})$. Since \(\bm{p}^{init}_{ij}\) is computed using fixed text embedding and prior from the pre-trained VLM, it may exhibit a distribution shift when applied to test data. To address this issue, we construct a cache that stores information from the previous \(i-1\) images to assist in prediction. The cache entries are defined as \(\bm{\mu}_m = (\bm{f}_m^{cache})\) for recognition tasks and \(\bm{\mu}_m = (\bm{f}_m^{cache}, \bm{b}_m^{cache})\) for detection tasks. Here, \(\bm{f}_m^{cache} \in \mathbb{R}^d\) represents the normalized mean of feature embeddings, while \(\bm{b}_m^{cache} = [w, h] \in \mathbb{R}^2\) aggregates the mean width and height of bounding boxes for similar spatial scales. These cached entries summarize historical information, enabling the model to adapt to the test data distribution dynamically. Importantly, the number of cached entries \(M\) is adaptive and can expand during inference to incorporate new knowledge observed during test time. The specific update mechanism for \(M\) and the cached entries will be detailed in subsequent sections.

As shown in Eq.~\eqref{Eq:3}, the evaluation of the likelihood \(P(\bm{x}_{ij}|\bm{\mu}_m)\) constitutes a critical step in deriving cache-based predictions, where \(\bm{x}_{ij}\) represents the current observation from the VLM and \(\bm{\mu}_m\) denotes the \(m\)-th cache entry. To compute the likelihood, we incorporate two key components: feature similarity, which measures the alignment between the visual embedding $\bm{f}_{ij}^v$ and the cached feature embedding $\bm{f}_{m}^{cache}$ for both recognition and detection tasks, and box similarity, which evaluates the spatial consistency between the bounding box $\bm{b}_{ij}$ and the cached spatial scale $\bm{b}_{m}^{cache}$ for detection task.

Consistent with many prior methods \cite{karmanov2024efficient,zhou2025bayesian}, the feature similarity is quantified as the cosine similarity between the current feature embedding \(\bm{f}_{ij^v}\) and the cached feature embedding \(\bm{f}_m^{cache}\), defined as:
\begin{equation}
S_{jm}^F = \cos(\bm{f}_{ij}^v, \bm{f}_m^{cache}).
\end{equation}
For box similarity, we extract $\bm{b}_{ij}[2:]$, which contains the width and height \([w, h]\), from the bounding box \(\bm{b}_{ij} = [x, y, w, h] \) and compare them with the cached scale \(\bm{b}_m^{cache}=[w, h]\). The scale similarity is computed as:
\begin{equation}
S_{jm}^B = 1 - \frac{\|{\bm{b}_{ij}[2:] - \bm{b}_m^{cache}}\|}{\sqrt{2}},
\end{equation}
where \(\|\cdot\|\) denotes the L2 norm and the difference is normalized by \(\sqrt{2}\). This normalization arises because the ranges of \(w\) and \(h\) are constrained to \([0, 1]\), resulting in a maximum difference of \(\|\bm{b}_{ij}[2:] - \bm{b}_m^{cache}\| = \sqrt{2}\) under perfect misalignment, thereby scaling the similarity score to the interval \([0, 1]\).

For recognition tasks, where no bounding box information is available, the box similarity term \({S}_{jm}^B\) is omitted, and the likelihood simplifies to:
\begin{equation}
\begin{aligned}
    P(\bm{x}_{ij}|\bm{\mu}_m) &\propto \text{exp}({S}_{jm}^F), \\
\end{aligned}
\label{Eq:61}
\end{equation}

For detection tasks, both terms are retained, ensuring that the likelihood reflects both semantic alignment and spatial consistency. This is achieved through a weighted combination of the box similarity \({S}_{jm}^B\) and the feature similarity \({S}_{jm}^F\), expressed as:
\begin{equation}
P(\bm{x}_{ij}|\bm{\mu}_m) \propto \text{exp}(w_s * {S}_{jm}^B + (1 - w_s) * {S}_{jm}^F),
\label{Eq:62}
\end{equation}
where \(w_s \in [0, 1]\) is a balance hyperparameter, \(\text{exp}(\cdot)\) denotes the exponential function.

Beyond the likelihood computation outlined in Eq.~\eqref{Eq:3}, it is imperative to consider the prior \(P(Y|\bm{\mu}_m)\) to fully characterize the cache-based prediction. Conventional test-time adaptation methods \cite{karmanov2024efficient, han2024dota} typically employ a fixed one-hot vector to represent this prior, assuming a static distribution across environments. However, this overlooks the dynamic nature of real-world scenarios; for instance, consider a person presenting with a fever—during the COVID-19 pandemic, the prior probability of COVID-19 should be significantly higher, whereas in a typical season, a common cold might be more probable. Therefore, we propose a dynamic prior by maintaining a set of vectors \( P(Y|\bm{\mu}_m) =\bm{v}^{cache}_m\), where \(\bm{v}^{cache}_m \in \mathbb{R}^K\) represents the probability distribution over \(K\) categories for each \(\bm{\mu}_m\). Unlike the fixed one-hot distribution used in previous methods, our approach adaptively learns priors through continuous updates of \(\bm{v}^{cache}_m\) based on the observed samples. The specific update mechanism for these priors will be detailed in section \ref{sec:update}.

To improve computational efficiency and scalability, we implement the cache-based prediction in matrix form. The cached feature embeddings are concatenated into a matrix \(\bm{F}_{cache} = [\bm{f}_1^{cache};\cdots;\bm{f}_M^{cache}] \in \mathbb{R}^{d \times M}\), the cached spatial scales into \(\bm{B}_{cache}= [\bm{b}_1^{cache};\cdots;\bm{b}_M^{cache}] \in \mathbb{R}^{2 \times M}\), and the cached priors into \(\bm{V}_{cache} = [\bm{v}_1^{cache};\cdots;\bm{v}_M^{cache}] \in \mathbb{R}^{K \times M}\). Based on Eq. \eqref{Eq:61}, Eq. \eqref{Eq:62} and Eq. \eqref{Eq:1}, the \(P(\bm{U}|\bm{x}_{ij}) = [P(\bm{\mu}_1|\bm{x}_{ij}), \cdots, P(\bm{\mu}_M|\bm{x}_{ij})] \in \mathbb{R}^M\) for object recognition is formulated as:
\begin{equation}
\begin{aligned}
    &P(\bm{U}|\bm{x}_{ij}) = \text{Softmax}({S}_{j}^F),\\
\end{aligned}
\label{Eq:71}
\end{equation}
and the \(P(\bm{U}|\bm{x}_{ij})\) for object detection is formulated as:
\begin{equation}
\begin{aligned}
    P(\bm{U}|\bm{x}_{ij}) &= \text{Softmax}(w_s * {S}_{j}^B + (1 - w_s) * {S}_{j}^F),
\end{aligned}
\label{Eq:72}
\end{equation}
where \(\text{Softmax}(\cdot)\) denotes the softmax function,
the feature similarity vector \({S}_j^F \in \mathbb{R}^M\) is computed as \({S}_j^F = \cos(\bm{f}_{ij}^v, \bm{F}_{cache})\), and the box similarity vector \({S}_j^B \in \mathbb{R}^M\) as \({S}_j^B = 1 - \frac{\|\bm{b}_{ij}[2:] - \bm{B}_{cache}\|_2}{\sqrt{2}}\), with \(\|\cdot\|_2\) denoting the L2 norm of matrix. Further, we derive the category-level posterior \(P(Y|\bm{x}_{ij})\) to capture the probabilistic distribution over \(K\) categories, which is formulated as: 
\begin{equation}
P(Y|\bm{x}_{ij}) = P(\bm{U}|\bm{x}_{ij}) *\bm{V}_{cache}^\top.
\label{Eq:8}
\end{equation}
This computed \(P(Y|\bm{x}_{ij})\) represents the cache-based probability \(\bm{p}^{cache}_{ij}\), integrating both adaptive likelihoods by updating \(\{\bm{\mu}_m\}_{i=1}^M\) and priors by updating \(\{\bm{v}^{cache}_m\}_{i=1}^M\).

\subsection{Uncertainty-Guided Fusion}
After processing through the cache, we compute a cache-based prediction \(\bm{p}^{cache}_{ij}\) for both recognition and detection tasks. This prediction leverages historical information stored in the cache, capturing both adaptive likelihoods and priors. To further enhance robustness, we fuse \(\bm{p}^{cache}_{ij}\) with the initial prediction \(\bm{p}^{init}_{ij}\) computed by the pre-trained VLM. The fusion is performed using an entropy-based uncertainty weighting mechanism, which dynamically balances the contributions of \(\bm{p}^{init}_{ij}\) and \(\bm{p}^{cache}_{ij}\) based on their respective uncertainties. The final prediction \(\bm{p}^{final}_{ij}\) is formulated as:
\begin{equation}
\bm{p}^{final}_{ij} = \frac{\exp(-\text{E}(\bm{p}^{init}_{ij})) \cdot \bm{p}^{init}_{ij} + \exp(-\text{E}(\bm{p}^{cache}_{ij})) \cdot \bm{p}^{cache}_{ij}}{\exp(-\text{E}(\bm{p}^{init}_{ij})) + \exp(-\text{E}(\bm{p}^{cache}_{ij}))},
\label{eq:final}
\end{equation}
where the Shannon entropy is defined as \(\text{E}(\bm{p}) = -\sum_{k=1}^K p_k \log p_k\). By combining \(\bm{p}^{init}_{ij}\) and \(\bm{p}^{cache}_{ij}\) in this manner, the resulting \(\bm{p}^{final}_{ij}\) achieves better robustness and generalization.

\subsection{Cache Adaptation}\label{sec:update}

The cache, consisting of \(\bm{F}_{cache}\), \(\bm{B}_{cache}\) (for detection tasks), and \(\bm{V}_{cache}\), plays a crucial role in supporting predictions for the test domain. To ensure the cache remains representative of the evolving test data distribution, it is dynamically updated as new test samples arrive. This continuous updating mechanism enables the cache to capture the characteristics of the test domain more accurately, thereby improving the model's ability to generate precise predictions for subsequent samples. Specifically, after predicting the current sample \(\bm{x}_i\), the cache is updated using the final prediction \(\bm{p}^{final}_{ij}\) derived from \(\bm{x}_i\). For recognition tasks, this takes the form of $\{(\bm{f}^v_{ij}, \bm{p}^{final}_{ij})\}_{j=1}^1$, while for detection tasks, it is $\{(\bm{f}^v_{ij}, \bm{b}_{ij}, \bm{p}^{final}_{ij})\}_{j=1}^N$. To ensure that only reliable information contributes to the cache, we filter out low-confidence proposals by retaining those that satisfy:
$
\max_k p^{final}_{ijk} \geq \tau_1,
$
where \(\tau_1\) is a confidence threshold. Only the remaining high-confidence proposals are used to update the cache.

For each retained proposal \(\bm{x}_{ij}\), we utilize its pre-computed matching distribution \(P(\bm{U}|\bm{x}_{ij})\), derived during prediction using Eq.~\eqref{Eq:71} or Eq.~\eqref{Eq:72}, to identify the most similar cache entry. Specifically, the matching score for each cache entry is defined as \(P(\bm{U}|\bm{x}_{ij})[m]\), and the best-matching entry is selected as:
\begin{equation}
m^* = \arg\max_{m} P(\bm{U}|\bm{x}_{ij})[m].
\label{eq:16}
\end{equation}

If \(P(\bm{U}|\bm{x}_{ij})[m^*] < \tau_2\), where \(\tau_2\) is a similarity threshold, or if the cache is still empty, we create a new cache entry by appending the current feature, spatial scale (for detection tasks only and thus omitted by recognition tasks), and class distribution:
\begin{equation}
\begin{aligned}
\bm{F}^{cache} &= \bm{F}^{cache} \oplus \bm{f}_{ij}^v, \\
\bm{B}^{cache} &= \bm{B}^{cache} \oplus \bm{b}_{ij}[2:],\\
\bm{V}^{cache} &= \bm{V}^{cache} \oplus \bm{p}^{final}_{ij}, \\
\bm{C}^{cache} &= \bm{C}^{cache} \oplus 1,
\end{aligned}
\label{eq:17}
\end{equation}
where \(\oplus\) denotes concatenation along the first dimension. Here, \(\bm{C}^{cache}\) tracks the number of updates for each cache entry, initialized as 1 for newly added entries.

Otherwise, if \(P(\bm{U}|\bm{x}_{ij})[m^*] \geq \tau_2\), indicating that a sufficiently similar cache entry exists, we update the matched entry \(\bm{\mu}_{m^*}\) using a statistical aggregation method. Specifically, for both recognition and detection tasks, we update the feature embeddings and priors. Additionally, for the detection task, we also update the spatial scales:
\begin{equation}
\begin{aligned}
\bm{f}_{m^*}^{cache} &= \frac{c_{m^*}^{cache} \cdot \bm{f}_{m^*}^{cache} + \bm{f}_{ij}^v}{c_{m^*}^{cache} + 1}, \\
\bm{b}_{m^*}^{cache} &= \frac{c_{m^*}^{cache} \cdot \bm{b}_{m^*}^{cache} + \bm{b}_{ij}[2:]}{c_{m^*}^{cache} + 1}, \\
\bm{v}_{m^*}^{cache} &= \frac{c_{m^*}^{cache} \cdot \bm{v}_{m^*}^{cache} + \bm{p}^{final}_{ij}}{c_{m^*}^{cache} + 1}, \\
c_{m^*}^{cache} &= c_{m^*}^{cache} + 1.
\end{aligned}
\label{eq:18}
\end{equation}

This updating mechanism ensures that each cache entry evolves in a confidence-aware and data-driven manner. This dynamic adaptation enables our method to progressively refine its internal model of the test domain without requiring access to training data or parameter updates.

In summary, we have presented BCA+, a unified Bayesian framework for test-time adaptation in both object recognition and detection, which is summarized in Algorithm \ref{alg:bca_plus}. By dynamically maintaining a cache of class embeddings, spatial scales, and adaptive priors, our method leverages historical information to refine predictions for incoming samples. The key innovation lies in its ability to seamlessly integrate the strengths of vision-language models with a robust, data-driven adaptation mechanism. Crucially, BCA+ is a {training-free} method that requires no backpropagation during inference. This design eliminates the computational bottleneck of gradient-based optimization, resulting in {significantly reduced inference time and memory usage}. In the next section, we will demonstrate that BCA+ not only achieves superior accuracy but also offers exceptional efficiency, making it a highly practical solution for real-world applications.

\begin{algorithm}[t]
  \caption{Bayesian Test-Time Adaptive Object Recognition and Detection (BCA+)}
  \label{alg:bca_plus}
  \textbf{Input}: A pre-trained VLM (CLIP or Grounding DINO), unlabeled test data $\{(\bm{x}_i)\}_{i=1}^{n}$, hyperparameters $\tau_1$, $\tau_2$, $w_s$\\
  \textbf{Procedure}: 
  \begin{algorithmic}[1]
  \STATE Initialize the cache: $\bm{F}^{cache} = \varnothing$; $\bm{V}^{cache} = \varnothing$; $\bm{C}^{cache} = \varnothing$; $\bm{B}^{cache} = \varnothing$ (for detection) ;
  \FOR{$i$ = 1:$n$}
    \STATE Process $\bm{x}_i$ with VLM to obtain initial outputs as $\{(\bm{f}_{ij}^v, \bm{p}^{init}_{ij})\}_{j=1}^1$ (recognition) or $\{(\bm{f}_{ij}^v, \bm{b}_{ij}, \bm{p}^{init}_{ij})\}_{j=1}^N$ (detection);
    % \STATE Compute cache-based predictions $\{\bm{p}^{cache}_{ij}\}_j$ using Eq.~\eqref{Eq:8}:
      \FOR{each proposal $\bm{x}_{ij}$}
        \STATE Calculate $P(\bm{U}|\bm{x}_{ij})$ using Eq.~\eqref{Eq:71} (recognition) or Eq.~\eqref{Eq:72} (detection);
        \STATE Calculate $\bm{p}^{cache}_{ij}$ using Eq. \eqref{Eq:8};
      % \ENDFOR
    \STATE Perform uncertainty-guided fusion based on $\bm{p}^{init}_{ij}$ and $\bm{p}^{cache}_{ij}$ to obtain $\bm{p}^{final}_{ij}$ using Eq. \eqref{eq:final};
      % \FOR{each proposal $\bm{x}_{ij}$}
        % \STATE $\bm{p}^{final}_{ij} = \frac{\exp(-\text{E}(\bm{p}^{init}_{ij})) \cdot \bm{p}^{init}_{ij} + \exp(-\text{E}(\bm{p}^{cache}_{ij})) \cdot \bm{p}^{cache}_{ij}}{\exp(-\text{E}(\bm{p}^{init}_{ij})) + \exp(-\text{E}(\bm{p}^{cache}_{ij}))}$;
      \ENDFOR
    % \STATE Update the cache using high-confidence proposals:
      \FOR{each $\bm{x}_{ij}$ where $\max_k p^{final}_{ijk} \geq \tau_1$}
        % \STATE Calculate matching distribution $P(\bm{U}|\bm{x}_{ij})$;
        \STATE Find best-matching $m^*$-th entry in the cache using Eq. \eqref{eq:16};
        \IF{$P(\bm{U}|\bm{x}_{ij})[m^*] < \tau_2$ or cache is empty}
          \STATE {Create new entry using Eq. \eqref{eq:17}};
        \ELSE
          \STATE {Update matched entry $\bm{\mu}_{m^*}$ using Eq. \eqref{eq:18}};
        \ENDIF
      \ENDFOR
    \STATE \textbf{return} final predictions $\{\bm{p}^{final}_{ij}\}_j$.
  \ENDFOR
  \end{algorithmic}
\end{algorithm}

\section{Experiments}\label{sec:exp}
\subsection{Experimental Setup}
\noindent\textbf{Datasets: } We evaluate our method BCA+ on standard benchmarks for both object recognition and object detection to demonstrate its effectiveness and robustness under various distribution shifts.

For object recognition, we use two established benchmarks: the Out-of-Distribution (OOD) benchmark and the Cross-Domain benchmark. The OOD benchmark evaluates robustness to data that significantly differs from the training distribution. It uses four challenging datasets derived from ImageNet \cite{deng2009imagenet}: ImageNet-A \cite{hendrycks2021natural}, ImageNet-V2 \cite{recht2019imagenet}, ImageNet-R \cite{hendrycks2021many}, and ImageNet-S \cite{wang2019learning}, which contain adversarial, re-sampled, and stylized images, respectively.
The Cross-Domain benchmark assesses a model's generalization across diverse image domains. It comprises ten datasets: Aircraft \cite{maji2013fine}, Caltech101 \cite{fei2004learning}, Cars \cite{krause20133d}, DTD \cite{cimpoi2014describing}, EuroSAT \cite{helber2019eurosat}, Flower102 \cite{nilsback2008automated}, Food101 \cite{bossard2014food}, Pets \cite{parkhi2012cats}, SUN397 \cite{xiao2010sun}, and UCF101 \cite{soomro2012ucf101}. Each dataset represents a distinct visual domain, providing a comprehensive evaluation of a model's adaptability to new tasks. 

For object detection, we evaluate our method on three datasets designed to simulate real-world distribution shifts through image corruption following previous works \cite{chen2023stfar,yoo2024and}. Specifically, we employ three corrupted datasets: (1) FoggyCityscapes \cite{sakaridis2018semantic}, a synthetic foggy version of Cityscapes \cite{cordts2016cityscapes}, where we test on the densest fog level ($\beta = 0.02$);(2) PASCAL-C, a corrupted version of the PASCAL VOC 2007 \cite{everingham2015pascal} test set (4,952 images) with 15 corruption types \cite{michaelis2019benchmarking} at severity 5; (3) COCO-C, which applies the same corruptions to 5,000 images from the MS-COCO 2017 \cite{lin2014microsoft} validation set.

\begin{table*}[t]
\begin{small}
\begin{center}
     \caption{Object Recognition Performance Comparison with the State-of-the-Art Methods on the OOD Benchmark. \\
     Metric: classification accuracy (\%);
     Bp-free: backpropagation-free at test time;
     Average: mean accuracy across all datasets;
     OOD Average: mean accuracy across four OOD datasets excluding ImageNet.}
     \label{tab:1}
\resizebox{\textwidth}{!}{
\begin{tabular}{l|cc|ccccc|cc}
\hline

Method &Venue &Bp-free  &ImageNet &ImageNet-A &ImageNet-V2 &ImageNet-R &ImageNet-S &Average &OOD Average  \\ \hline
\multicolumn{10}{c}{\textbf{Visual Backbone: ResNet-50}}  \\ \hline
TPT \cite{shu2022test}         & NIPS22  & \XSolidBrush  &60.74 &26.67 &54.70 &59.11 &35.09 &47.26& 43.89 \\
DiffTPT \cite{feng2023diverse}  & ICCV23  & \XSolidBrush  &60.80 &\bf31.06 &55.80 &58.80& 37.10 &48.71 &45.69 \\
C-TPT \cite{yoonc}  & ICLR24  & \XSolidBrush  &61.2   &25.6 &54.8 &59.7 &35.7 &47.4   &44.0 \\
TDA \cite{karmanov2024efficient}  & CVPR24  & \CheckmarkBold  &61.35 &30.29 &55.54 &62.58 &38.12 &49.58 &46.63\\
O-TPT \cite{sharifdeen2025tpt}&CVPR25&\XSolidBrush &58.97&23.07 &53.11 &54.47 &33.98&44.72 &41.16\\
FreeTTA \cite{dai2025free}&CVPR25&\CheckmarkBold &61.51 &30.67 &55.89 &63.02 &37.94 &49.81 &46.88\\
BCA \cite{zhou2025bayesian} & CVPR25  &\CheckmarkBold   & 61.81  & 30.35& 56.58  &62.89 &38.08 &49.94   &46.98 \\ \hline
CLIP \cite{radford2021learning} &ICML21  & \CheckmarkBold   &59.81 &23.24 &52.91 &60.72 &35.48 &46.43 &43.09  \\ 
CLIP+LA          & Ours &\CheckmarkBold&61.32 & 28.93 & 56.01  &62.28 & 37.72&49.25&46.24\\
BCA+ & Ours &\CheckmarkBold & \bf62.25  & 30.68 &\bf56.81  &\bf63.42 &\bf39.47&\bf50.53&\bf47.60 \\
\hline
\multicolumn{10}{c}{\textbf{Visual Backbone: ViT-B/16}}  \\ \hline
TPT \cite{shu2022test}         & NIPS22  & \XSolidBrush  &68.98 &54.77 &63.45 &77.06 &47.94 &62.44 &60.81 \\
DiffTPT \cite{feng2023diverse}  & ICCV23  & \XSolidBrush  &70.30 &55.68 &65.10 &75.00 &46.80 &62.28 &60.52 \\
C-TPT \cite{yoonc}  & ICLR24  & \XSolidBrush  & 69.3  & 52.9 &63.4 &78.0 &48.5&62.4 &60.7 \\
TDA \cite{karmanov2024efficient}  & CVPR24  & \CheckmarkBold  & 69.51& 60.11& 64.67& 80.24 &50.54 &65.01 &63.89 \\
MTA \cite{zanella2024test}  &  CVPR24 &  \CheckmarkBold &70.08 &58.06 &64.24 &78.33 &49.61 &64.06 & 62.56\\
PromptAlign \cite{samadhalign} &  NIPS24 & \XSolidBrush & -  &59.37 &65.29 &79.33 &50.23 & -&63.55 \\
Zero \cite{farina2024frustratingly} & NIPS24 & \CheckmarkBold&69.06& 61.35& 64.13& 77.28& 48.29& 64.02  &62.76 \\ 
O-TPT \cite{sharifdeen2025tpt}&CVPR25&\XSolidBrush &67.33 &49.87 &61.65 &72.55 &47.12 &59.71 &57.80\\
FreeTTA \cite{dai2025free}&CVPR25&\CheckmarkBold &70.21 &61.41 &64.92 &80.49 &50.88 &65.58 &64.42\\
BCA \cite{zhou2025bayesian} & CVPR25  &\CheckmarkBold   & 70.22  & 61.14 &  64.90 &80.72 &50.87 &65.37   &64.16 \\ \hline
CLIP \cite{radford2021learning} &ICML21   & \CheckmarkBold &  68.34 &49.89&61.88 &77.65 &48.24 &61.20 &59.42  \\ 
CLIP+LA          & Ours &\CheckmarkBold& 70.08  &59.22 &  64.10 &79.89 &50.23&64.70&63.36\\
BCA+ & Ours &\CheckmarkBold & \bf71.16  & \bf61.44 &\bf65.47 &\bf81.22 &\bf51.32&\bf66.12&\bf64.86  \\
\hline
 \hline
\end{tabular}}
\end{center}
\end{small}
\end{table*}

\begin{table*}[t]
\caption{Object Recognition Performance Comparison with the State-of-the-Art Methods on Cross Domain benchmark.\\ 
     Metric: classification accuracy (\%).}
\resizebox{\textwidth}{!}{
\begin{tabular}{l|cccccccccc|c}
\hline

Method   &Aircraft &Caltech101 &Cars &DTD &EuroSAT &Flower102 &Food101 &Pets &SUN397 &UCF101 &Average \\ 
\hline
\multicolumn{12}{c}{\textbf{Visual Backbone: ResNet-50}}  \\ \hline
TPT \cite{shu2022test} &17.58 &87.02 &58.46 &40.84 &28.33 &62.69 &74.88 &84.49 &61.46 &60.82 &57.66  \\
DiffTPT \cite{feng2023diverse}&17.60 &86.89 &60.71 &40.72 &41.04 &63.53 &79.21 &83.40 &62.72 &62.67 &59.85   \\
C-TPT \cite{yoonc} & 17.5 &87.4   & 57.3  &43.1 & 29.4  &65.3   & 76.0  &84.0 & 62.1  & 60.7  & 58.3  \\
TDA \cite{karmanov2024efficient} &17.61 &89.70 &57.78 &43.74 &42.11 &\bf68.74 &77.75 &86.18 &62.53 &\bf64.18 &61.03   \\
HisTPT\cite{zhang2024historical}&18.1  & 87.2  & \bf61.3  & 41.3& 42.5  & 67.6  & \bf81.3  &84.9 & 63.5  & 64.1  &61.2   \\
FreeTTA \cite{dai2025free} &17.83 &90.12 &58.01 &44.21 &\bf43.64 &68.26 &77.98 &86.44 &62.84 &63.97 &61.33\\
BCA\cite{zhou2025bayesian} & 19.89 & 89.70  & 58.13  & 48.58 & 42.12  & 66.30 &77.19   &85.58  &63.38   &63.51   & 61.44\\ \hline
CLIP \cite{radford2021learning} &16.11 &87.26 &55.89 &40.37 &25.79 &62.77 &74.82 &82.97 &60.85 &59.48 &56.63 \\
CLIP+LA & 19.02 & 89.24 &57.64   &45.65 &  36.74 &66.07   &76.59   &84.63 & 63.29  & 61.15&60.00 \\
BCA+ & \bf20.72  & \bf90.65   & 59.99 & \bf49.07 & 42.65  & 68.55 &77.84   &\bf86.59  &\bf64.73   &64.17& \bf62.50 \\\hline
\multicolumn{12}{c}{\textbf{Visual Backbone: ViT-B/16}}  \\ \hline
TPT \cite{shu2022test} &24.78 &94.16 &66.87 &47.75 &42.44 &68.98 &84.67 &87.79 &65.50 &68.04 &65.10 \\
DiffTPT \cite{feng2023diverse}       &25.60 &92.49 &67.01 &47.00 &43.13 &70.10 &87.23 &88.22 &65.74 &62.67 &65.47\\
C-TPT \cite{yoonc} &23.9   & 94.1  & 66.7  & 46.8 &48.7   &69.9   &84.5   &87.4 & 66.0  &66.7   &    65.5\\
TDA \cite{karmanov2024efficient} &23.91 &94.24 &67.28 &47.40 &58.00 &71.42 &86.14 &88.63 &67.62 &70.66 &67.53\\
MTA \cite{zanella2024test} &25.20   &94.21   &68.47   &45.90  &45.36   &68.06   &85.00   &88.24 &66.67   &68.69   &   65.58\\
PromptAlign \cite{samadhalign}& 24.80  & 94.01  & 68.50  & 47.24 & 47.86  &  72.39 & \bf86.65  & 90.76 & 67.54  & 69.47  &   66.92\\
HisTPT\cite{zhang2024historical}& 26.9  & 94.5  & \bf69.2  & 48.9 & 49.7  &  71.2 &  89.3 & 89.1 &  67.2 &  70.1 &   67.6\\ 
Zero \cite{farina2024frustratingly}& 24.42  & 94.14  &  68.48 & 45.86 & 43.77  & 66.82  & 84.58  &87.20 & 66.90  & 68.57  &   65.07\\ 
FreeTTA \cite{dai2025free} &25.11 &94.63 &67.34 &46.96 &\bf62.93 &71.62 &86.62 &90.11 &67.76 &\bf71.16 &68.42\\
BCA \cite{zhou2025bayesian} &28.59  &94.69   &66.86   & 53.49& 56.63  &73.12   &85.97   &90.43 &68.41   &67.59   & 68.59\\ \hline
CLIP \cite{radford2021learning}  &23.22 &93.55 &66.11 &45.04 &50.42 &66.99 &82.86 &86.92 &65.63 &65.16 &64.59 \\ 
CLIP+LA &28.36 & 94.22 &66.31 &53.12 &54.98 &73.07 &85.86 &90.14 &68.07 &67.40&68.15\\
BCA+ & \bf28.79&\bf94.81&67.38&\bf54.29&55.59&\bf74.45&86.36&\bf91.12&\bf69.77&68.04&\bf69.06 \\
\hline
\end{tabular}}
\label{tab:2}
\end{table*}

\noindent\textbf{{Implementation details:}} 
Our experiment is conducted on the Pytorch platform. The batch size is set to 1 for both tasks to suit the application scenario of test time adaptation.

For object recognition, following \cite{zhou2025bayesian}, we adopt pre-trained CLIP models with ResNet-50 \cite{he2016deep} and ViT-B/16 \cite{dosovitskiy2020image} as the visual encoder, and a Transformer \cite{vaswani2017attention} as the text encoder. The hyperparameters $\tau_1$ and $\tau_2$ are both set to 0.8.

For object detection, we use pre-trained Grounding DINO models with Swin-T \cite{liu2021swin} and Swin-B \cite{liu2021swin} as the visual encoder, and BERT \cite{devlin2019bert} as the text backbone. The hyperparameters $\tau_1$ and $\tau_2$ are also set to 0.8, and the balance weight $w_s$ is set to 0.2.

\noindent\textbf{Comparison Methods:} We compare our method against a range of state-of-the-art test-time adaptation approaches, categorized based on the task and methodology.

For object recognition, we focus on comparing with existing TTA methods built upon the CLIP framework, as it is the dominant paradigm now. These methods can be broadly divided into two categories:
1. Training-based methods: These approaches adapt the model during testing by performing backpropagation, which is computationally expensive, such as TPT \cite{shu2022test}, DiffTPT \cite{feng2023diverse}, C-TPT \cite{yoonc}, PromptAlign \cite{samadhalign}, HisTPT \cite{zhang2024historical} and O-TPT \cite{sharifdeen2025tpt}.
2. Training-free methods: These methods avoid backpropagation entirely, relying on memory or statistical updates for real-time adaptation, such as TDA \cite{karmanov2024efficient}, MTA \cite{zanella2024test}, Zero \cite{farina2024frustratingly}, FreeTTA \cite{dai2025free} and BCA \cite{zhou2025bayesian}.

For object detection, there are currently no existing TTA methods designed for open-vocabulary detectors like Grounding DINO. Therefore, we compare against state-of-the-art TTA methods that are built on traditional, closed-set detectors such as Faster R-CNN, including SHOT \cite{liang2020we}, T3A \cite{iwasawa2021test}, Self-Training \cite{xu2021end}, TTAC \cite{su2022revisiting}, STFAR \cite{chen2023stfar} and W3TTAOD \cite{yoo2024and}. To further validate the effectiveness of BCA+, we extend BCA \cite{zhou2025bayesian} and TDA \cite{karmanov2024efficient} from the CLIP-based recognition setting to the Grounding DINO-based detection task for a direct comparison with same backbone.

\subsection{Comparisons with State-of-the-art}
For object recognition, we follow the prompt ensembling strategy of \cite{karmanov2024efficient,zhou2025bayesian}, using multiple context prompt templates to generate $K$ initial class embeddings. We provide a detailed comparison of our method: CLIP \cite{radford2021learning} serves as the baseline; CLIP+LA denotes the variant that incorporates likelihood adaptation (LA) on top of the baseline; and BCA+ represents our full model, which integrates both likelihood and prior adaptation.

\noindent\textbf{Results on the OOD benchmark.}
We first evaluate our method on the Out-of-Distribution (OOD) benchmark, which assesses a model's robustness to data that differs significantly from its training set. The results, presented in Table~\ref{tab:1}, demonstrate the superior performance of our BCA+ method.
With the ResNet-50 backbone, BCA+ achieves an average accuracy of 50.53\% and an OOD average of 47.60\%, setting a new state-of-the-art. It surpasses the previous best training-free method, BCA \cite{zhou2025bayesian}, by 0.59\% in average accuracy and 0.62\% in OOD average. Notably, BCA+ even exceeds the performance of the backpropagation-based DiffTPT \cite{feng2023diverse} by 1.82\% in average accuracy. The strong performance of CLIP+LA (49.25\% average) confirms the effectiveness of likelihood adaptation. However, the further improvement to BCA+ highlights the critical contribution of prior adaptation.
The trend is consistent with the more powerful ViT-B/16 backbone. BCA+ achieves the best results across all metrics, with an average accuracy of 66.12\% and an OOD average of 64.86\%. It sets a new state-of-the-art, improving upon BCA by 0.75\% in average accuracy and 0.70\% in OOD average. BCA+ also outperforms other strong baselines like TDA and Zero. These results conclusively show that BCA+ achieves the highest robustness on the OOD benchmark, validating the effectiveness of our unified adaptation framework.

\noindent\textbf{Results on the Cross-Domain benchmark.}
We further evaluate our method on the Cross-Domain (CD) benchmark, which assesses a model's generalization across ten diverse object recognition tasks. The results, presented in Table~\ref{tab:2}, demonstrate the consistent superiority of our BCA+ method.
With the ResNet-50 backbone, BCA+ achieves an average accuracy of 62.50\%, outperforming all competing methods. It surpasses the previous state-of-the-art BCA \cite{zhou2025bayesian} by 1.06\% and the strong training-free baseline TDA \cite{karmanov2024efficient} by 1.47\%. Notably, BCA+ sets new records on eight out of the ten datasets, including a significant 2.83\% improvement on Aircraft and a 0.49\% improvement on DTD over BCA. The performance of CLIP+LA (60.00\% average) confirms the benefit of likelihood adaptation, while the further gain to BCA+ underscores the critical role of prior adaptation in handling diverse domains.
The trend is even more pronounced with the ViT-B/16 backbone. BCA+ achieves a new state-of-the-art average accuracy of 69.06\%, surpassing the previous best (BCA \cite{zhou2025bayesian}) by 0.47\%. BCA+ achieves the highest accuracy on nine out of ten datasets, including a new record of 28.79\% on Aircraft. These results conclusively show that BCA+ achieves the best cross-domain generalization, validating its robustness across a wide range of visual tasks.

\begin{table*}[htbp]
\centering
\caption{Object Detection Performance Comparison with the state-of-the-art methods on FoggyCityscapes.\\
Metric: Average Precision@50(\%);
Bp-free: backpropagation-free at test time.}
\label{tab:FoggyCityscapes}
\begin{tabular}{l|ccc|c c c c c c c c|c}
\hline
Methods &Venue &Framework &Bp-free&Pson & Rder & Car & Tuck & Bus & Train & Mcle & Bcle & mAP$_{50}$  \\
\hline
\multicolumn{13}{c}{\textbf{Visual Backbone: ResNet-50 }}  \\ \hline
SHOT \cite{liang2020we} &ICML20&Faster RCNN& \XSolidBrush& 26.7 & 30.3 & 36.9 & 16.8 & 28.9 & 6.4 & 14.3 & 23.3 & 23.0 \\
T3A \cite{iwasawa2021test} &NIPS21&Faster RCNN& \XSolidBrush& 22.6 & 23.0 & 31.9 & 7.7 & 14.8 & 1.0 & 7.9 & 19.7 & 16.6 \\
Self-Training \cite{xu2021end} &CVPR21&Faster RCNN& \XSolidBrush& 27.7 & 30.8 & 41.4 & 12.8 & 27.4 & 4.2 & 14.8 & 26.1 & 23.1 \\
TTAC \cite{su2022revisiting} &NIPS22&Faster RCNN& \XSolidBrush& 24.5 & 27.3 & 33.4 & 14.6 & 26.1 & 5.8 & 14.1 & 21.5 & 20.9 \\
STFAR\cite{chen2023stfar}  &ARXIV23&Faster RCNN& \XSolidBrush& 28.8 & 32.0 & 42.4 & 15.1 & 30.1 & 11.2 & 15.5 & 26.2 & 25.1 \\
\hline
\multicolumn{13}{c}{\textbf{Visual Backbone: Swin-T}}  \\ \hline
TDA \cite{karmanov2024efficient} &CVPR24&Grounding DINO& \CheckmarkBold&35.27&4.87&46.43&22.82&32.30&0.38&28.68&29.86&25.08\\
{BCA} \cite{zhou2025bayesian} &CVPR25&Grounding DINO& \CheckmarkBold & \bf41.01&\bf5.61&\bf47.85&21.86&33.40&\bf0.91&28.47&29.82&26.12\\\hline
GDINO \cite{liu2024grounding}&ECCV24&Grounding DINO& \CheckmarkBold &30.10&3.42&46.26&22.41&31.98&0.08&27.96&28.87&23.88 \\
GDINO+LA&Ours&Grounding DINO& \CheckmarkBold&38.18&3.82&46.51&\bf22.71&33.03&0.60&28.95&30.27&25.51 \\
BCA+&Ours&Grounding DINO& \CheckmarkBold&39.03&3.61&46.98&22.26&\bf35.26&0.87&\bf29.37&\bf35.87&\bf26.65  \\
\hline
\multicolumn{13}{c}{\textbf{Visual Backbone: Swin-B}}  \\ \hline
{TDA} \cite{karmanov2024efficient} &CVPR24&Grounding DINO& \CheckmarkBold 
&\bf38.29&25.53&50.09&30.49&\bf45.92&11.98&33.12&39.06&34.31
\\
BCA \cite{zhou2025bayesian} &CVPR25&GroundingDINO& \CheckmarkBold&36.53&24.42&50.77&\bf30.83&45.25&\bf20.29&33.20&\bf42.49&35.47
\\\hline
GDINO \cite{liu2024grounding} &ECCV24&Grounding DINO& \CheckmarkBold&34.95&20.35&51.56&30.01&45.72&0.87&\bf35.21&32.06&31.34  \\
GDINO+LA &Ours&Grounding DINO& \CheckmarkBold
&35.88&22.30&51.48&30.52&45.17&16.52&33.57&41.50&33.49
\\
BCA+ &Ours&Grounding DINO& \CheckmarkBold
&38.15&\bf26.30&\bf52.30&30.77&45.66&\bf20.29&33.89&42.41&\bf36.22
\\\hline
\end{tabular}
\end{table*}

For object detection, we use the category name as the input text prompt for Grounding DINO to perform open-vocabulary detection. We provide a detailed ablation study of our method: Grounding DINO \cite{liu2024grounding} serves as the baseline; GDINO+LA denotes the variant that incorporates only likelihood adaptation (LA), which updates the class embeddings and spatial scales; and BCA+ represents our full model, which integrates both likelihood and prior adaptation for comprehensive test-time adaptation.

\noindent\textbf{Results on the FoggyCityscapes dataset.}
We evaluate our method on the FoggyCityscapes dataset, which simulates dense foggy conditions and poses a significant challenge for object detection. The results, presented in Table~\ref{tab:FoggyCityscapes}, demonstrate the superior performance of our BCA+ method.
For the Swin-T backbone, BCA+ achieves an mAP$_{50}$ of 26.65\%, outperforming the baseline GDINO (23.88\%) and the training-free method TDA (25.08\%). It also surpasses our ablation model GDINO+LA (25.51\%), which only uses likelihood adaptation, highlighting the benefit of incorporating prior adaptation. BCA+ sets new records on challenging categories like Bicycle (35.87\%) and Bus (35.26\%).
The performance gain is even more significant with the more powerful Swin-B backbone. BCA+ achieves an mAP$_{50}$ of 36.22\%, significantly outperforming all baselines. It surpasses the previous state-of-the-art BCA \cite{zhou2025bayesian} by 0.75\% and the strong baseline TDA by 1.91\%. BCA+ achieves the best results on most categories, including Person, Car, and Train. These results conclusively show that BCA+ is highly effective for test-time adaptation in object detection under adverse weather conditions.

\begingroup
\setlength{\tabcolsep}{4pt}  
\begin{table*}[htbp]
\centering
\caption{Object Detection Performance Comparison with the state-of-the-art methods on PASCAL-C.\\
Metric: mean Average Precision@50(\%).}
\label{tab:pascalc}
\begin{tabular}{l|c c c c c c c c c c c c c c c|c}
\hline
{Methods} & Brit & Contr & Defoc & Elast & Fog & Frost & Gauss & Glass & Impul & Jpeg & Motn & Pixel & Shot & Snow & Zoom & Average \\
\hline
\multicolumn{17}{c}{\textbf{Visual Backbone: ResNet-50 }}  \\ \hline
SHOT \cite{liang2020we} & 72.0 & 31.7 & 18.9 & 46.6 & 67.5 & 45.8 & 12.0 & 11.6 & 16.4 & 41.8 & 19.7 & 33.1 & 19.9 & 42.5 & 27.6 & 33.8 \\
T3A \cite{iwasawa2021test} & 36.9 & 12.5 & 11.0 & 19.7 & 32.7 & 20.6 & 6.1 & 6.4 & 6.5 & 14.8 & 10.1 & 13.2 & 8.4 & 16.8 & 13.8 & 15.3 \\
Self-Training \cite{xu2021end} & 67.9 & 39.3 & 2.6 & 52.5 & 65.7 & 47.2 & 11.9 & 20.2 & 12.1 & 29.3 & 4.1 & 6.9 & 17.4 & 44.9 & 9.5 & 28.8 \\
TTAC \cite{su2022revisiting} & 72.2 & 40.4 & 29.3 & 58.1 & 68.7 & 50.4 & 29.8 & 28.7 & 33.6 & 46.4 & 29.2 & 46.1 & 35.1 & 48.0 & 34.9 & 43.4 \\
STFAR \cite{chen2023stfar} & 67.3 & 51.8 & 34.8 & 55.7 & 65.2 & 50.7 & 32.4 & 34.6 & 36.3 & 49.4 & 34.6 & 55.7 & 37.8 & 50.9 & 34.8 & 46.1 \\
\hline
\multicolumn{17}{c}{\textbf{Visual Backbone: Swin-T}}  \\ \hline
TDA \cite{karmanov2024efficient} &64.61&47.71&37.33&32.41&61.77&52.19&27.05&20.48&29.35&41.45&28.50&14.15&31.85&48.07&24.35&37.42\\
{BCA} \cite{zhou2025bayesian} &68.11&49.42&38.46&34.42&64.27&53.21&28.37&22.24&31.90&42.81&30.39&14.81&\bf33.68&49.37&24.97&39.10 \\
\hline
GDINO \cite{liu2024grounding} &63.04&45.66&35.58&31.33&60.92&50.52&25.82&18.84&27.99&39.34&26.99&13.39&30.49&45.89&23.38&35.95  \\
GDINO+LA
& 67.41&48.50&38.50&34.43&63.88&54.72&27.53&21.90&30.95&45.89&30.33&14.40&32.89&48.37&24.66&38.96\\
BCA+ 
&\bf70.25&\bf50.72&\bf40.79&\bf37.08&\bf67.83&\bf56.38&\bf28.59&\bf22.48&\bf31.74&\bf47.40&\bf30.75&\bf14.87&33.66&\bf51.87&\bf25.65&\bf40.67 \\
\hline
\multicolumn{17}{c}{\textbf{Visual Backbone: Swin-B}}  \\ \hline

TDA \cite{karmanov2024efficient} &86.28&74.33&57.98&58.45&85.98&76.41&57.82&43.84&60.80&74.31&57.91&69.45&63.79&78.36&38.84&65.64
\\
{BCA} \cite{zhou2025bayesian} &88.66&75.73&59.84&61.00&88.02&77.69&60.21&45.98&62.46&76.32&60.55&71.22&64.89&80.67&40.66   &67.59
\\\hline
GDINO  \cite{liu2024grounding}
&85.16&70.69&56.12&55.56&84.56&73.75&54.94&41.17&57.72&71.44&56.08&66.17&60.54&76.48&36.94&63.15
\\
GDINO+LA 
&88.26&75.25&60.16&60.17&87.21&78.16&59.30&45.47&62.41&75.89&59.75&71.14&65.16&80.16&39.58 &67.20
\\
BCA+
&\bf89.39&\bf77.85&\bf62.04&\bf62.71&\bf88.87&\bf79.94&\bf61.81&\bf47.67&\bf64.87&\bf78.12&\bf61.81&\bf73.51&\bf67.28&\bf82.42&\bf41.42&\bf69.31
\\
\hline
\end{tabular}
\end{table*}
\endgroup

\noindent\textbf{Results on the PASCAL-C dataset.}
We evaluate our method on the PASCAL-C dataset, which applies 15 different types of corruptions to the PASCAL VOC 2007 test set. The results, presented in Table~\ref{tab:pascalc}, demonstrate the strong performance of our BCA+ method.
With the Swin-T backbone, BCA+ achieves an average mAP$_{50}$ of 40.67\%, significantly outperforming the baseline GDINO (35.95\%) and the training-free method TDA (37.42\%). It also surpasses GDINO+LA (38.96\%), which only uses likelihood adaptation, by 1.71\%, highlighting the benefit of incorporating prior adaptation. BCA+ sets new state-of-the-art results on 13 out of 15 corruption types, with notable gains on challenging corruptions like "Brightness" (+7.21) and "Elastic Transform" (+2.65).
The performance gain is even more substantial with the Swin-B backbone. BCA+ achieves an average mAP$_{50}$ of 69.31\%, setting a new state-of-the-art and surpassing the previous best (BCA \cite{zhou2025bayesian}) by 1.72\%. BCA+ achieves the best results on all 15 corruption types, demonstrating its exceptional robustness. These results conclusively show that BCA+ is highly effective for test-time adaptation in object detection under diverse distribution shifts.

\begingroup
\setlength{\tabcolsep}{4pt}  
\begin{table*}[t]
\centering
\caption{Object Detection Performance Comparison with the state-of-the-art methods on COCO-C.\\
Metric: mean Average Precision@50(\%).}
\label{tab:cococ}
\begin{tabular}{l|c c c c c c c c c c c c c c c|c}
\hline
{Methods} & Brit & Contr & Defoc & Elast & Fog & Frost & Gauss & Glass & Impul & Jpeg & Motn & Pixel & Shot & Snow & Zoom & Average \\\hline
\multicolumn{17}{c}{\textbf{Visual Backbone: ResNet-50}}  \\ \hline
SHOT \cite{liang2020we} & 40.9 & 26.6 & 14.7 & 19.7 & 41.5 & 26.7 & 11.0 & 7.2 & 12.1 & 16.4 & 11.0 & 9.7 & 13.0 & 22.0 & 6.4 & 18.6 \\
T3A \cite{iwasawa2021test} & 28.8 & 15.9 & 8.3 & 11.3 & 28.9 & 17.2 & 4.6 & 3.1 & 5.2 & 9.0 & 5.8 & 4.1 & 5.8 & 13.8 & 3.5 & 11.0 \\
Self-Training \cite{xu2021end}& 38.1 & 28.4 & 14.7 & 25.5 & 38.5 & 27.9 & 16.7 & 11.4 & 18.8 & 23.8 & 16.0 & 24.5 & 18.6 & 27.6 & 7.8 & 22.6 \\
TTAC \cite{su2022revisiting} & 38.3 & 29.5 & 15.1 & 28.2 & 39.0 & 28.5 & 16.8 & 14.3 & 18.0 & 23.2 & 14.3 & 24.8 & 19.3 & 26.7 & 8.7 & 23.0 \\
STFAR \cite{chen2023stfar} & 39.1 & 31.1 & 16.8 & 29.0 & 39.0 & 29.2 & 19.2 & 15.4 & 20.1 & 26.1 & 17.2 & 28.3 & 21.0 & 29.5 & 10.2 & 24.7 \\
W3TTAOD \cite{yoo2024and} & 36.4 & 27.2 & 14.0 & 27.2 & 37.4 & 27.2 & 13.6 & 13.6 & 16.1 & 22.3 & 14.2 & 22.2 & 16.6 & 23.7 & 8.3 & 21.3 \\
\hline

\multicolumn{17}{c}{\textbf{Visual Backbone: Swin-T}}  \\ \hline
TDA \cite{karmanov2024efficient} &45.53 & 24.43 & 20.33 & 27.08 & 45.91 & 32.31 & 15.77 & 9.93 & 16.72 & 26.25 & 16.85 & \bf8.99 & 17.10 & 28.22 & 9.62&23.00\\
{BCA} \cite{zhou2025bayesian} &46.34 & 25.18 & 21.03 & 27.45 & 46.74 & 33.92 & 15.97 & 10.91 & 17.33 & 26.19 & 16.05 & 8.89 & 17.97 & 29.02 & 9.87& 23.52  \\\hline
GDINO \cite{liu2024grounding} &42.91&23.83&19.45&23.76&43.26&30.36&15.03&10.22&15.87&23.77&16.56&7.31&16.99&25.51&9.35& 21.61 \\
GDINO+LA&46.05 & 25.53 & 20.93 & 26.79 & 45.27 & 32.43 & 16.21 & 10.90 & 18.00 & 26.32 & 18.74 & 7.02 & 17.52 & 28.28 & 9.82 &23.32\\
BCA+ 
&\bf49.62&\bf26.56&\bf22.20&\bf28.65&\bf49.79&\bf35.65&\bf17.97&\bf11.47&\bf18.67&\bf28.38&\bf19.10&7.83&\bf19.86&\bf30.10&\bf10.08&\bf25.06\\\hline
\multicolumn{17}{c}{\textbf{Visual Backbone: Swin-B}}  \\ \hline
TDA \cite{karmanov2024efficient} 
&56.86&42.26&30.35&38.15&58.21&46.72&32.28&22.64&32.84&42.12&29.94&38.97&34.17&44.14&14.89&37.64
\\
{BCA} \cite{zhou2025bayesian} 
&57.21&42.32&31.00&38.90&58.84&46.43&32.17&23.37&33.53&42.49&31.15&39.56&35.76&45.42&14.92&38.20
\\\hline
GDINO \cite{liu2024grounding} 
&55.36&40.40&29.11&36.51&56.26&44.39&30.27&21.38&31.30&40.26&28.78&36.68&32.67&42.71&13.44&35.97
\\
GDINO+LA 
&57.96&42.38&31.12&37.85&58.93&45.76&31.94&23.70&33.92&42.91&30.88&38.71&34.58&44.65&15.04&38.02
\\
BCA+ 
&\bf60.34&\bf44.99&\bf32.25&\bf40.85&\bf60.92&\bf49.06&\bf34.04&\bf24.42&\bf35.33&\bf44.91&\bf31.72&\bf41.17&\bf36.59&\bf47.85&\bf15.24&\bf39.98
\\
\hline
\end{tabular}
\end{table*}
\endgroup

\noindent\textbf{Results on the COCO-C dataset.}
We further evaluate our method on the COCO-C dataset, which applies 15 different types of corruptions to the MS-COCO validation set, providing a comprehensive test of robustness. The results, presented in Table~\ref{tab:cococ}, demonstrate the consistent superiority of our BCA+ method across all corruption types.
With the Swin-T backbone, BCA+ achieves an average mAP$_{50}$ of 25.06\%, significantly outperforming the baseline GDINO (21.61\%) and the training-free method TDA (23.00\%). It also surpasses GDINO+LA (23.32\%), which only uses likelihood adaptation, by a large margin of 1.74\%, highlighting the substantial benefit of incorporating prior adaptation. BCA+ sets new state-of-the-art results on all 15 corruption types, with particularly large gains on challenging corruptions like "Brightness" (+3.57), "Defocus Blur" (+1.45), and "Fog" (+4.52).
The performance gain is even more pronounced with the more powerful Swin-B backbone. BCA+ achieves an average mAP$_{50}$ of 39.98\%, setting a new state-of-the-art and surpassing the previous best (BCA \cite{zhou2025bayesian}) by 1.78\%. BCA+ achieves the best results on all 15 corruption types, demonstrating its exceptional robustness. These results conclusively show that BCA+ is highly effective for test-time adaptation in object detection under a wide variety of distribution shifts.

\subsection{Ablation Studies}

\begin{table}[h]
\begin{small}
\begin{center}
     \caption{Efficiency and performance comparison on the {\it ImageNet} dataset (Object Recognition). Visual Backbone: ResNet-50; GPU: RTX 4070 Ti SUPER GPU.}
     \label{tab:efficiency_recognition}
\begin{tabular}{lcccc}
\hline
Method  & Time (min) & Memory (M) & Accuracy (\%)  \\ \hline
CLIP \cite{radford2021learning}    &2.23 &753 &59.81 \\ \hline
TPT \cite{shu2022test} &572.13 &21396 & 60.74  \\
TDA \cite{karmanov2024efficient} &11.93 & 1174& 61.35  \\
{BCA+} & 2.89 & 805 & 62.25  \\
\hline
  \end{tabular}
\end{center}
\end{small}
\end{table}

\begin{table}[h]
\begin{small}
\begin{center}
     \caption{Efficiency and performance comparison on the {\it COCO-C-Brit} dataset (Object Detection). Visual Backbone: Swin-T; GPU: RTX 4070 Ti SUPER GPU.}
     \label{tab:efficiency_detection}
\begin{tabular}{lccc}
\hline
Method  &  Time (min) & Memory (M) & mAP$_{50}$ (\%)  \\ \hline
GDINO \cite{karmanov2024efficient} &14.5 &3025 &42.91 \\ \hline 
TDA \cite{karmanov2024efficient} &42.6&4586&45.53\\
{BCA+} &19.8 &3862&49.62\\
\hline
  \end{tabular}
\end{center}
\end{small}
\end{table}

\noindent{\bf Efficiency and performance comparison.}
A key advantage of our training-free framework is its high practicality, seamlessly combining superior performance with exceptional computational efficiency. To demonstrate this, we conduct a comprehensive comparison of efficiency (inference time and memory usage) and effectiveness (accuracy) on both object recognition and object detection tasks.
For object recognition on the ImageNet dataset (Table~\ref{tab:efficiency_recognition}), our BCA+ method achieves the highest accuracy of 62.25\%, outperforming the baseline CLIP by 2.44\% and the previous state-of-the-art TDA by 0.90\%. Crucially, BCA+ achieves this superior performance with exceptional efficiency: its inference time (2.89 minutes) is drastically lower than the backpropagation-based TPT (572.13 minutes) and significantly faster than the training-free TDA (11.93 minutes). Its memory usage (805M) is also minimal, comparable to the baseline CLIP (753M) and far lower than TPT (21396M).
For object detection on the COCO-C-Brit dataset (Table~\ref{tab:efficiency_detection}), the results show a similar trend. BCA+ achieves the best mAP$_{50}$ of 49.62\%, surpassing both the baseline Grounding DINO (42.91\%) and the TTA method TDA (45.53\%). Most importantly, BCA+ maintains its efficiency advantage: it processes the dataset in 17.8 minutes, which is significantly faster than TDA (72.6 minutes) and only marginally slower than the baseline Grounding DINO (14.5 minutes). Its memory footprint is also reasonable, being lower than TDA's.
In summary, BCA+ excels not only in accuracy but also in efficiency. It consistently outperforms existing methods by achieving higher performance with significantly less inference time and memory usage, making it a highly practical solution for real-world, real-time applications.

\begin{table}[h]
\begin{small}
\begin{center}
     \caption{Ablation Study on Different Cache Update Strategies.}
     \label{tab:update_strategies}
\begin{tabular}{l|ccc}
\hline
Dataset & Count-based & Momentum-based & Delayed \\ \hline
ImageNet & 62.25 & 61.80 & 60.92 \\ 
FoggyCityscapes & 26.65 & 26.29 & 25.83 \\ 
\hline
\end{tabular}
\end{center}
\end{small}
\end{table}

\noindent{\bf Analysis of update strategies.}
The effectiveness of our cache-based adaptation relies on the strategy used to update the cached entries. In our main method, we use a count-based averaging strategy, where each cached component (feature embedding, spatial scale, prior) is updated as a running average weighted by a counter $c_m$ that tracks the number of updates: $\bm{v}_m^{new} = \frac{c_m \cdot \bm{v}_m^{old} + \bm{v}_{new}}{c_m + 1}$. This allows the cache to adapt quickly in the early stages and stabilize over time.
To validate the design of our update mechanism, we compare it against two alternative strategies:
1.  Momentum-based Update: This strategy uses a fixed momentum coefficient $\alpha=0.95$ to perform an exponential moving average: $\bm{v}_m^{new} = \alpha \cdot \bm{v}_m^{old} + (1 - \alpha) \cdot \bm{v}_{new}$.
2.  Delayed Update: This strategy only updates a cache entry after it has been matched $k$ times. We set $k=5$, meaning an entry is only updated on its 5th, 10th, 15th, ... match.
We evaluate these three strategies on the ImageNet dataset for object recognition and the FoggyCityscapes dataset for object detection. As shown in Table~\ref{tab:update_strategies}, the count-based method achieves the best performance on both tasks. The momentum-based method performs reasonably well but is less adaptive to rapid changes in the test distribution. The delayed update method performs the worst, as it significantly slows down the adaptation process and fails to incorporate new information in a timely manner.
These results demonstrate that our count-based update strategy is crucial for achieving high performance. It provides a natural and robust way to balance the influence of new observations with historical knowledge, making our method highly effective for dynamic environments.

\begin{figure*}[t]
    \centering
    \includegraphics[width=\textwidth]{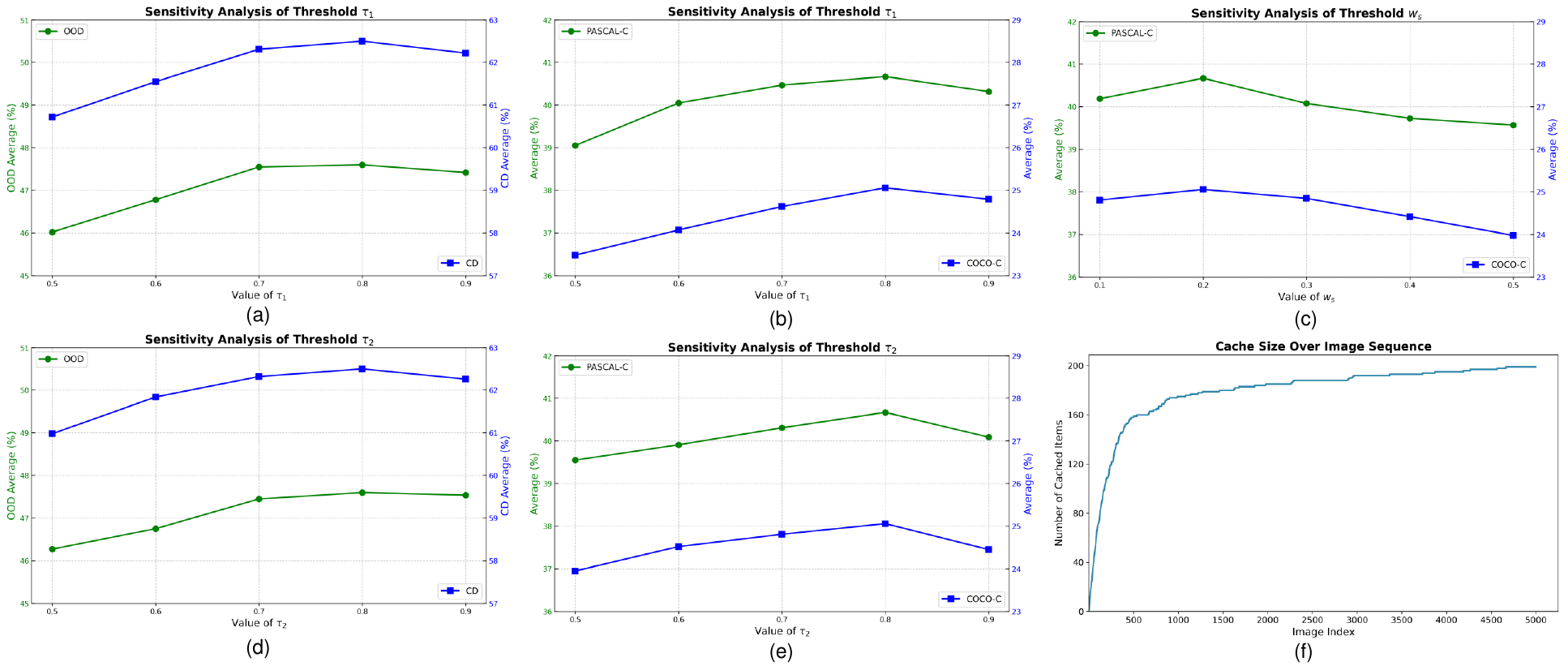}
    \caption{Hyperparameter sensitivity analysis and cache dynamics analysis. (a) and (d) show the sensitivity of object recognition performance to the confidence threshold $\tau_1$ and the similarity threshold $\tau_2$ on the OOD and Cross-Domain benchmarks, respectively. (b) and (e) show the sensitivity of object detection performance to $\tau_1$ and $\tau_2$ on the PASCAL-C and COCO-C datasets, respectively. (c) shows the sensitivity of detection performance to the balance weight $w_s$ on the PASCAL-C and COCO-C datasets. (f) shows the number of cached entries $M$ over the image sequence on the COCO-C-Brit dataset.}
    \label{fig:sensitivity}
\end{figure*}

\begin{figure*}[t]
    \centering
    \includegraphics[width=\textwidth]{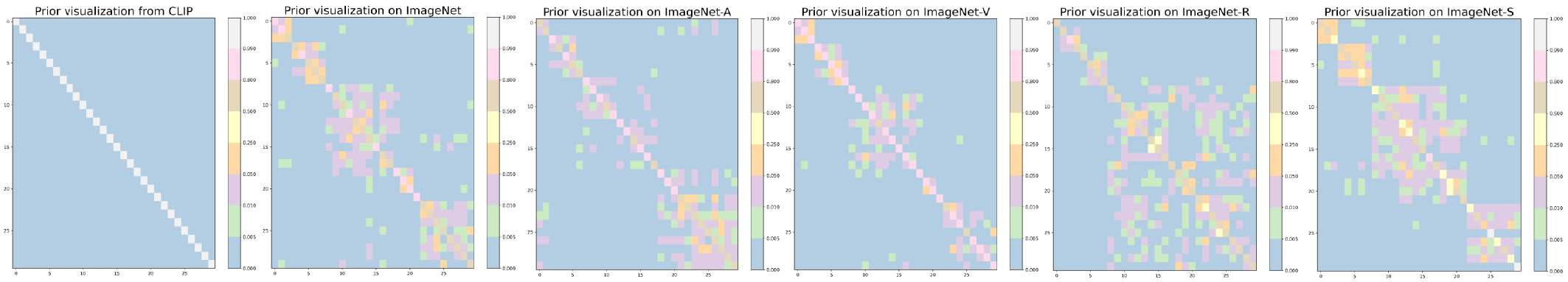}
    \caption{Prior visualization on OOD benchmark.}
    \label{fig:prior_vis}
\end{figure*}

\noindent{\bf Parameter sensitivity analysis.}
To understand the influence of key hyperparameters on our method's performance, we conduct a comprehensive sensitivity analysis, the results of which are presented in Figure~\ref{fig:sensitivity}(a)-(e). The analysis is performed using the CLIP model with a ResNet-50 visual backbone for object recognition and the Grounding DINO model with a Swin-T visual backbone for object detection.
For object recognition, Figure~\ref{fig:sensitivity}(a) and \ref{fig:sensitivity}(d) show the impact of the confidence threshold $\tau_1$ and the similarity threshold $\tau_2$ on the OOD and Cross-Domain (CD) benchmarks, respectively. The performance is relatively stable across a wide range of $\tau_1$ values (0.6-0.9), with a peak around 0.8. Similarly, the model is robust to variations in $\tau_2$, achieving optimal performance when $\tau_2$ is between 0.7 and 0.9, with a peak at 0.8.
For object detection, Figure~\ref{fig:sensitivity}(b) and \ref{fig:sensitivity}(e) present the sensitivity to $\tau_1$ and $\tau_2$ on the PASCAL-C and COCO-C datasets. The trends are consistent with the recognition task, demonstrating that our method is not overly sensitive to the exact choice of these thresholds. The optimal performance is consistently achieved when both $\tau_1$ and $\tau_2$ are set to 0.8. Figure~\ref{fig:sensitivity}(c) shows the impact of the balance weight $w_s$, which controls the fusion of feature and scale similarity. The performance is stable when $w_s$ is between 0.1 and 0.3, with an optimal value of 0.2, indicating that scale information provides a valuable but secondary signal to feature similarity.
Overall, these results demonstrate that our method is robust to the choice of hyperparameters. The optimal values ($\tau_1=0.8$, $\tau_2=0.8$, $w_s=0.2$) are effective across different tasks and datasets, making BCA+ practical and easy to deploy.

\noindent{\bf Cache Size over Image Sequence.}
To visualize the dynamic nature of our adaptation process, we analyze the growth of the cache size $M$ as test images arrive sequentially. Figure~\ref{fig:sensitivity}(f) plots the number of cached entries $M$ against the number of processed images on the COCO-C-Brit dataset. Initially, $M$ grows rapidly as the model encounters novel object categories and creates new cache entries. As more images are processed, the rate of growth slows down, indicating that the cache has captured the majority of the common categories in the test domain. The final cache size converges to a value close to, but not necessarily equal to, the total number of categories $K$ (represented by the dashed line), as some categories may be too rare to meet the creation threshold. This dynamic evolution demonstrates that our cache adapts organically to the test data distribution, expanding to learn new concepts and stabilizing as knowledge is accumulated.

\begin{figure*}[t]
    \centering
    \includegraphics[width=0.95\textwidth]{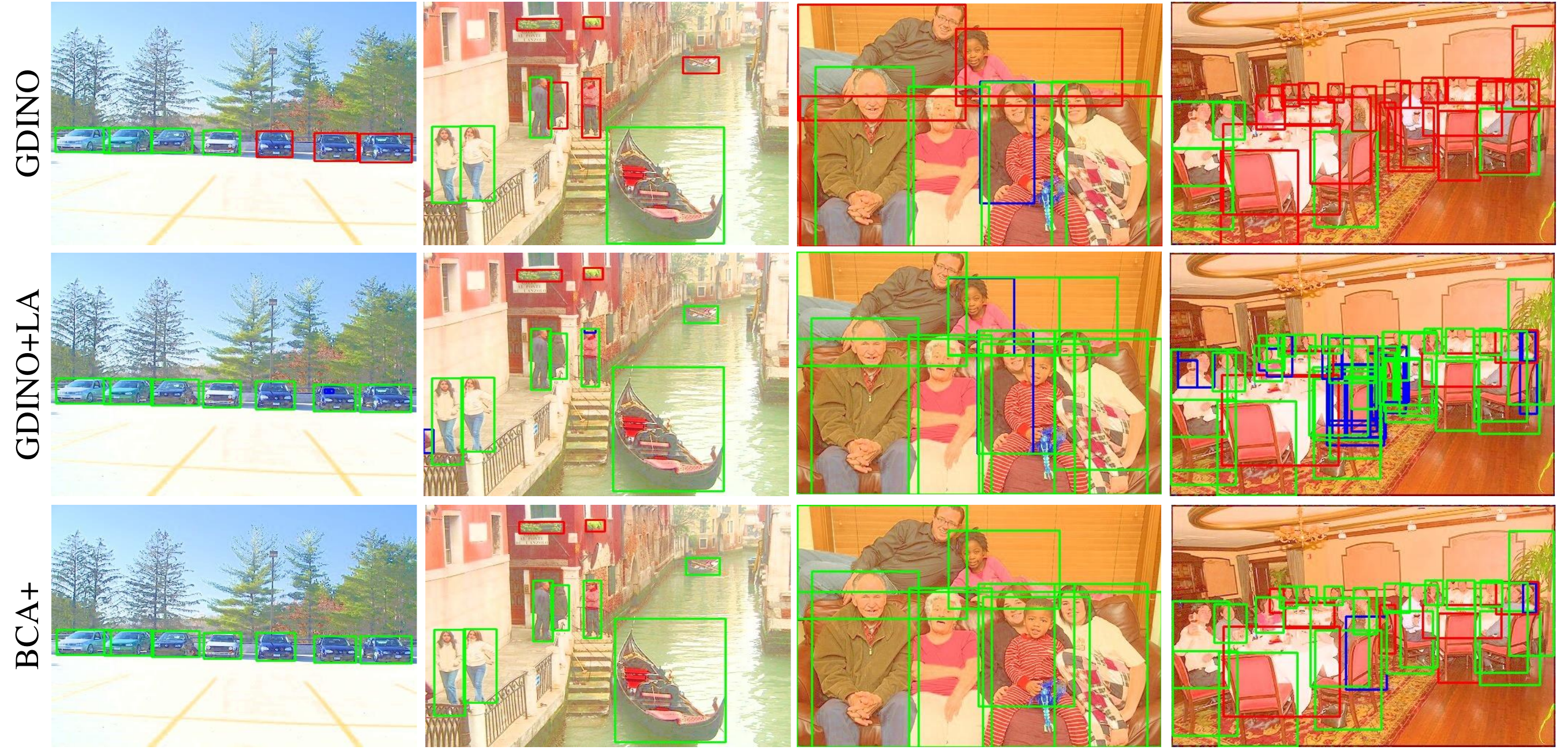}
    \caption{ Qualitative results on PASCAL-C-Brit. Green, red and blue boxes represent true positives, false negatives and false positives, respectively. }
    \label{fig:detections}
\end{figure*}

\noindent{\bf Prior visualization.} 
To provide an intuitive understanding of the dynamic prior adaptation in our BCA+ framework, we visualize the learned class priors on the OOD benchmark for object recognition. We use the CLIP model with a ViT-B/16 visual backbone and compare the priors learned by BCA+ with the fixed, one-hot priors from the original CLIP model.
Specifically, we select the first 30 categories from the OOD dataset. For each category, we select a cache entry based on its pseudo-label and extract its learned prior distribution $P(Y|\bm{\mu}_m)$. The results are shown in Figure~\ref{fig:prior_vis}. It is evident that the priors learned by BCA+ are significantly different from the fixed priors of CLIP and vary dramatically across different OOD datasets. This visualization clearly demonstrates that the test data distribution deviates from the pre-training assumption, and thus, a static prior is suboptimal. The ability of BCA+ to dynamically adjust the prior according to the current test environment is crucial for robust performance.

\noindent{\bf Qualitative Results on Object Detection.}
To provide a visual comparison of the performance, we present qualitative results on the PASCAL-C-Brit dataset using the Grounding DINO model with a Swin-T visual backbone. As shown in Figure~\ref{fig:detections}, we compare the detection outputs of three methods: the Baseline (the original Grounding DINO), Baseline+LA (with only likelihood adaptation), and our full BCA+ model.
The visualizations demonstrate that our BCA+ method produces the most accurate and robust detections under severe corruption. The baseline model often suffers from false positives (FP), marked in blue, and missed detections (FN), marked in red. Baseline+LA shows improved robustness by reducing some false alarms, but it still struggles with challenging cases. In contrast, BCA+ achieves the best performance, successfully detecting objects with higher precision and recall. This is attributed to the synergistic effect of our dual adaptation mechanism, where the dynamic prior helps to suppress unlikely categories (reducing blue FP) and the refined likelihood ensures accurate localization (reducing red FN). As a result, BCA+ generates the most true positives (TP), marked in green, while maintaining the lowest number of FP and FN. These qualitative results visually confirm the quantitative superiority of BCA+ observed in our experiments.

\begin{table}[h]
\begin{small}
\begin{center}
     \caption{Ablation Study on the Effectiveness of Uncertainty-Guided Fusion.}
     \label{tab:fusion_ablation_combined}
\begin{tabular}{l|cccc}
\hline
Method & OOD & CD & PASCAL-C & COCO-C\\ \hline
Baseline & 43.09 & 56.63 & 35.95 & 21.61  \\
Cache  & 41.80 & 55.55 & 34.20 & 20.85\\
Average & 46.83 & 59.75 & 38.25 & 23.05\\
BCA+ & {47.60} & {62.50} & {40.67} & {25.06}\\
\hline
\end{tabular}
\end{center}
\end{small}
\end{table}

\noindent{\bf Performance of Individual Components and Fusion.}
To analyze the contribution of each component in our framework, we conduct an ablation study on the effectiveness of the uncertainty-guided fusion mechanism. The results, presented in Table~\ref{tab:fusion_ablation_combined}, are obtained using the CLIP model with a ResNet-50 visual backbone for object recognition and the Grounding DINO model with a Swin-T visual backbone for object detection.
We evaluate four different strategies: (1) the Baseline, which uses only the initial prediction from the pre-trained VLM (CLIP or Grounding DINO); (2) Cache, which uses only the cache-based prediction $\bm{p}^{cache}_{ij}$. As described in Section~\ref{sec:method}, the cache is initialized as empty and is dynamically updated with high-confidence final predictions during inference. For the first 15\% of the test sequence, the cache is still being populated, we use baseline prediction for evaluation. For the remaining 85\%, we use the pure cache prediction for evaluation; (3) Average, which fuses the baseline and cache predictions with a simple arithmetic mean; and (4) BCA+, which uses our proposed entropy-based uncertainty-guided fusion.
The results show that the cache-based prediction alone (Cache) performs worse than the baseline on all benchmarks, indicating that the raw cache prediction is not sufficient and that the initial VLM prediction contains vital information. However, when combined with the baseline, the performance improves significantly. The simple averaging strategy (Average) boosts the performance, demonstrating the benefit of combining historical information with the initial VLM prediction. Our full BCA+ model, which uses uncertainty-guided fusion, achieves the best performance on all tasks. It outperforms the simple averaging strategy by a clear margin, highlighting the importance of intelligently weighting the two predictions based on their confidence. This analysis validates that our uncertainty-guided fusion strategy is a crucial component for achieving superior performance.

\begin{table}[h]
\begin{small}
\begin{center}
     \caption{Performance Comparison for the last 50\% samples.}
     \label{tab:last50}
\begin{tabular}{l|cccc}
\hline
Method & OOD & CD & PASCAL-C & COCO-C\\ \hline
Baseline &43.12&56.92&35.63&21.52  \\
Baseline+LA&46.88&60.47&39.54&24.18\\
BCA+ &49.12&64.07&42.21&26.46\\
\hline
\end{tabular}
\end{center}
\end{small}
\end{table}

\noindent{\bf{Performance comparison for the last 50\% samples.}} 
To understand how well different methods adapt to new environments over time, we analyze their performance on the last 50\% of the test sequence. This setup helps us understand how well these methods can adapt to new environments over time. By this point, the cache in adaptive methods has been sufficiently populated with historical information, allowing us to evaluate the mature stage of the adaptation process. The results, presented in Table~\ref{tab:last50}, are obtained using the CLIP model with a ResNet-50 visual backbone for object recognition and the Grounding DINO model with a Swin-T visual backbone for object detection.
We compare three variants: the Baseline (the pre-trained VLM without adaptation), Baseline+LA (the baseline with only likelihood adaptation), and our full BCA+ model. The results show a significant performance gap between the methods on the later samples. BCA+ achieves the highest accuracy on all benchmarks, outperforming the baseline by a large margin and also surpassing Baseline+LA. For instance, on the OOD benchmark, BCA+ achieves 49.12\% accuracy, which is 2.24\% higher than Baseline+LA. This demonstrates that the combination of likelihood and prior adaptation in BCA+ enables the model to learn a more robust and accurate representation of the test domain over time, leading to superior performance in the later stages of inference.

\noindent{\bf Component analysis.} 
To validate the effectiveness of each component in our proposed BCA+ framework, we conduct a comprehensive ablation study on both object recognition and object detection tasks. We compare three variants: the baseline model (CLIP or Grounding DINO), the baseline with only likelihood adaptation (LA), and the full BCA+ model with likelihood adaptation and prior adaptation (LA+PA). The results are presented in Table~\ref{tab:1} to Table~\ref{tab:cococ}.
Based on the experimental results, we can draw the following conclusions:
1. Both Likelihood and Prior Adaptation are Beneficial: By comparing the baseline model with the model that incorporates only likelihood adaptation (LA), we observe a clear performance gain. This confirms that updating class embeddings (and spatial scales for detection) based on incoming data helps the model adapt to the new environment. Crucially, we also find that incorporating prior adaptation (PA) alone provides a significant improvement. This demonstrates that dynamically updating the model's belief about class frequencies is a powerful mechanism for handling distribution shifts.
2. Likelihood and Prior Adaptation are Synergistic: The full BCA+ model (LA+PA) consistently outperforms both the baseline and the model with only LA. This performance gap indicates a synergistic effect: the combination of likelihood and prior adaptation is more effective than the sum of its parts. The updated prior guides the model to focus on more probable categories, while the refined likelihood ensures accurate feature matching, leading to superior overall performance.
3. Prior Adaptation is Critical and Often Overlooked: The performance gain from adding prior adaptation is substantial. For instance, on the ImageNet dataset (Table~\ref{tab:1}), BCA+ (LA+PA) achieves a 1.08\% higher accuracy than the model with only LA. This highlights that prior adaptation is not just an auxiliary component, but a crucial factor for robust test-time adaptation. Critically, this finding underscores a key limitation of many existing TTA methods, which focus solely on likelihood adaptation (e.g., updating class embeddings) while ignoring the dynamic updating of the prior. By explicitly modeling and updating the prior, BCA+ provides a more complete and effective solution to the problem of distribution shift.

\section{Conclusion}\label{sec:conclusion}

In this paper, we present Bayesian Class Adaptation plus BCA+, a unified, training-free framework for test-time adaptation in both object recognition and object detection. By leveraging the power of vision-language models like CLIP and Grounding DINO, BCA+ addresses the challenge of distribution shifts during inference without requiring backpropagation or access to the original training data.
Our method introduces a dynamic cache that simultaneously adapts both the likelihood (class embeddings and spatial scales) and the prior (class distribution) based on high-confidence predictions from incoming test samples. This dual adaptation mechanism, formulated within a Bayesian inference framework, allows BCA+ to correct both the model's semantic understanding and its contextual confidence. The final prediction is generated by fusing the initial VLM prediction with a cache-based prediction using an uncertainty-guided strategy, which intelligently weights their contributions based on confidence.
Extensive experiments on standard benchmarks for both object recognition and object detection demonstrate that BCA+ achieves state-of-the-art performance. It significantly outperforms existing test-time adaptation (TTA) methods while maintaining high inference efficiency, making it a highly practical solution for real-world applications. The ablation studies validate the effectiveness of each component, particularly highlighting the critical role of prior adaptation, which is often overlooked in existing work.
In conclusion, BCA+ provides a robust, efficient, and unified solution for enhancing the adaptability of VLMs to dynamic real-world environments. Our work underscores the importance of a comprehensive Bayesian approach to test-time adaptation, paving the way for more intelligent and resilient vision systems.

\section*{Acknowledgments}
{This work was supported in part by Chinese National Natural Science Foundation Projects U23B2054, 62276254, 62306313 and 62276048, the Beijing Science and Technology Plan Project Z231100005923033, Beijing Natural Science Foundation L221013, and the InnoHK program.}


\bibliographystyle{IEEEtran}
\bibliography{cite}

\begin{IEEEbiographynophoto}
%[{\includegraphics[width=1in,height=1.25in,clip,keepaspectratio]{bios/Lihua.pdf}}]
{Lihua Zhou} received the Ph.D. degree from the University of Electronic Science and Technology of China in 2024. He is currently a Postdoctoral Research Fellow at the Centre for Artificial Intelligence and Robotics (CAIR), Hong Kong Institute of Science and Innovation, Chinese Academy of Sciences, China, Hong Kong. His research interests include machine learning, computer vision, and transfer learning.

\end{IEEEbiographynophoto}

\begin{IEEEbiographynophoto}
{Mao Ye} received the B.S. degree from Sichuan Normal University in mathematics, Chengdu, China, in 1995, the M.S. degree in mathematics from the University of Electronic Science and Technology of China, Chengdu, in 1998, and the Ph.D. degree in mathematics from the Chinese University of Hong Kong, China, in 2002. He has been a short-term Visiting Scholar with the University of Queensland and the University of Pennsylvania. He is currently a Professor and the Director of CVLab with the University of Electronic Science and Technology of China, Chengdu. His research interests include machine learning and computer vision. In these areas, he has published over 90 papers in leading international journals or conference proceedings. He has served on the editorial board of Engineering Applications of Artificial Intelligence. He was a Co-Recipient of the Best Student Paper Award at the IEEE ICME 2017.

\end{IEEEbiographynophoto}

\begin{IEEEbiographynophoto}
{Shuaifeng Li} received the B.S. degree in computer science and technology from the University of Electronic Science and Technology of China, Chengdu, China, in 2020. He is currently pursuing the M.S. degree and Ph.D. degree at the University of Electronic Science and Technology of China, Chengdu, China. His current research interests include computer vision and transfer learning.

\end{IEEEbiographynophoto}

\begin{IEEEbiographynophoto}
{Nianxin Li} received the B.S. degree in Computer Science and Technology from the University of Electronic Science and Technology of China, Chengdu, China, in 2022. He is currently pursuing the M.S. degree and Ph.D. degree at the University of Electronic Science and Technology of China, Chengdu, China. His current research interests include machine learning, computer vision, and transfer learning.

\end{IEEEbiographynophoto}

\begin{IEEEbiographynophoto}
{Jinlin Wu} received the B.S. degree from the University of Electronic Science and Technology of China, Chengdu, China in 2017, and the Ph.D. degree from the Institute of Automation, Chinese academy of science, Chinese Academy of Sciences in 2022. He is currently an Assistant Professor at the Centre for Artificial Intelligence and Robotics (CAIR), Hong Kong Institute of Science and Innovation, Chinese Academy of Sciences. His research interests include medical image analysis, video understanding, and multimodal large language models.
\end{IEEEbiographynophoto}

\begin{IEEEbiographynophoto}
{Xiatian Zhu} received the Ph.D. from Queen Mary University of London. He won the Sullivan Doctoral Thesis Prize 2016, an annual award representing the best doctoral thesis submitted to a U.K. University in computer vision. His research interests include computer vision and machine learning.

\end{IEEEbiographynophoto}

\begin{IEEEbiographynophoto}
{Lei Deng} received the B.Eng. degree from the Department of Electronic Engineering, Shanghai Jiao Tong University, Shanghai, China, in 2012, and the Ph.D. degree from the Department of Information Engineering, The Chinese University of Hong Kong, Hong Kong, in 2017. In 2015, he was a Visiting Scholar with the School of Electrical and Computer Engineering, Purdue University, West Lafayette, IN, USA. He is currently an Assistant Professor with the College of Electronics and Information Engineering, Shenzhen University, Shenzhen, China. His current research interest includes real-time communication (RTC).
\end{IEEEbiographynophoto}

\begin{IEEEbiographynophoto}
{Hongbin Liu} is a Professor with the Institute of Automation (IA), Chinese Academy of Sciences (CAS), Beijing, China, an Executive Deputy Director of the Centre for Artificial Intelligence and Robotics (CAIR), Hong Kong Institute of Science and Innovation, Chinese Academy of Sciences, Hong Kong. Dr. Liu is also an Adjunct Reader and the Director of the Haptic Mechatronics and Medical Robotics (HaMMeR) Laboratory, School of Biomedical Engineering and Imaging Sciences, King’s College London (KCL), London, U.K. His group has been focusing on research and development of medical robotic systems with advanced haptic perception and interaction capabilities, to enable safer and more effective minimally invasive diagnosis and treatment for patients. His research has led to the clinical translation of a series of flexible robotic endoscopic systems for applications such as colonoscopy, bronchoscopy, and vascular surgeries.

\end{IEEEbiographynophoto}

\begin{IEEEbiographynophoto}
{Jiebo Luo} is the Albert Arendt Hopeman Professor of engineering and a Professor of computer science with the University of Rochester, Rochester, NY, USA, which he joined after a prolific career of fifteen years with Kodak Research Laboratories in 2011. He has authored over 600 technical papers and holds more than 90 U.S. patents. His research interests include computer vision, NLP, machine learning, data mining, multimedia, computational social science, and digital health. Prof. Luo has been involved in numerous technical conferences, including serving as Program Co-Chair of ACM Multimedia 2010, IEEE CVPR 2012, ACM ICMR 2016, and IEEE ICIP 2017, and General Co-Chair of ACM Multimedia 2018, and IEEE ICME 2024. He served on the editorial boards of the IEEE Transactions on Pattern Analysis and Machine Intelligence, IEEE Transactions on Multimedia, IEEE Transactions on Circuits and Systems for Video Technology, IEEE Transactions on Big Data, ACM Transactions on Intelligent Systems and Technology, Pattern Recognition, and Intelligent Medicine. He was the Editor-in-Chief of IEEE Transactions on Multimedia from 2020 to 2022. Professor Luo is also a Fellow of ACM, AAAI, SPIE, and IAPR, and a Member of Academia Europaea and the US National Academy of Inventors.

\end{IEEEbiographynophoto}

\begin{IEEEbiographynophoto}
{Zhen Lei} received the B.S. degree in automation from the University of Science and Technology of China, in 2005, and the Ph.D. degree from the Institute of Automation, Chinese Academy of Sciences, in 2010. He is currently a Professor with the Institute of Automation, Chinese Academy of Sciences. He has published over 200 papers in international journals and conferences with more than 30000 citations in Google Scholar and an H-index of 81. His research interests are in computer vision, pattern recognition, image processing, and face recognition in particular. He is an IAPR Fellow and an AAIA Fellow. He was a winner of the 2019 IAPR Young Biometrics Investigator Award. He was the Program Co-Chair of IJCB2023, the Competition Co-Chair of IJCB2022, and the area chair of several conferences. He is an Associate Editor of the IEEE Transactions on Information Forensics and Security, IEEE Transactions on Biometrics, Behavior, and Identity Science, Pattern Recognition, Neurocomputing, and IET Computer Vision.

\end{IEEEbiographynophoto}

\vfill

\end{document}